\documentclass[10pt,twocolumn,letterpaper]{article}

\usepackage[]{cvpr}

\usepackage{microtype}

\newcommand{\ours}{Gnomon4D}
\newcommand{\trk}{GeoDelta}
\newcommand{\humanego}{EgoManip-5000H}

\usepackage{colortbl}

\definecolor{cvprblue}{rgb}{0.21,0.49,0.74}
\usepackage[pagebackref,breaklinks,colorlinks,allcolors=cvprblue]{hyperref}

\usepackage{graphicx}
\usepackage{subcaption}
\usepackage{tikz}

\usepackage[dvipsnames]{xcolor}
\usepackage{wrapfig}
\usepackage{float,wrapfig}
\usepackage{subcaption}
\usepackage{algpseudocode} 
\usepackage{algorithm}
\usepackage{verbatim}

\definecolor{citypink}{RGB}{227, 108, 194}
\definecolor{cityblue}{RGB}{128, 159, 225}

\usepackage{indentfirst}
\usepackage{titletoc}
\usepackage{cuted}
\usepackage{stfloats}
\usepackage{capt-of}
\usepackage{textcomp, gensymb}
\usepackage{courier}
\usepackage{multirow}

\usepackage{url}
\usepackage{balance}

\def\confName{CVPR}
\def\confYear{2027}

\title{Gnomon4D: Learning Where and When to Focus for Geometric Motion-Grounded World-Action Modeling}

\author{
    Lingwei Dang$^{1,2,3*}$\hspace{0.5em}%
    Yan Zhou$^{3}$\hspace{0.5em}%
    Juntong Li$^{1}$\hspace{0.5em}%
    Zhenhao Zhang$^{2,3*}$\hspace{0.5em}%
    Yiteng Chen$^{1}$\hspace{0.5em}%
    Xiaoqiang Liu$^{3\dagger}$
     \\ \vspace{1mm}
    Liang An$^{4}$\hspace{0.5em}%
    Hongwen Zhang$^{5}$\hspace{0.5em}%
    Wei Min$^{6}$\hspace{0.5em}%
    Guoxin Zhang$^{3}$\hspace{0.5em}%
    Pengfei Wan$^{3}$\hspace{0.5em}%
    Yebin Liu$^{2}$\hspace{0.5em}%
    Qingyao Wu$^{1\ddagger}$
    \\ \vspace{1mm}
    $^{1}$ South China University of Technology
    \quad
    $^{2}$ Tsinghua University
    \quad
    $^{3}$ KlingAI Research
    \\
    $^{4}$ Nanjing University
    \quad
    $^{5}$ Beijing Normal University
    \quad
    $^{6}$ Shadow AI
    \\ \vspace{1mm}
    {\href{https://droliven.github.io/Gnomon4D}{\textbf{\texttt{https://droliven.github.io/Gnomon4D}}}}
}

\begin{document}

\twocolumn[{
  \renewcommand\twocolumn[1][]{#1}
  \maketitle
  \begin{center}
  \includegraphics[width=\textwidth]{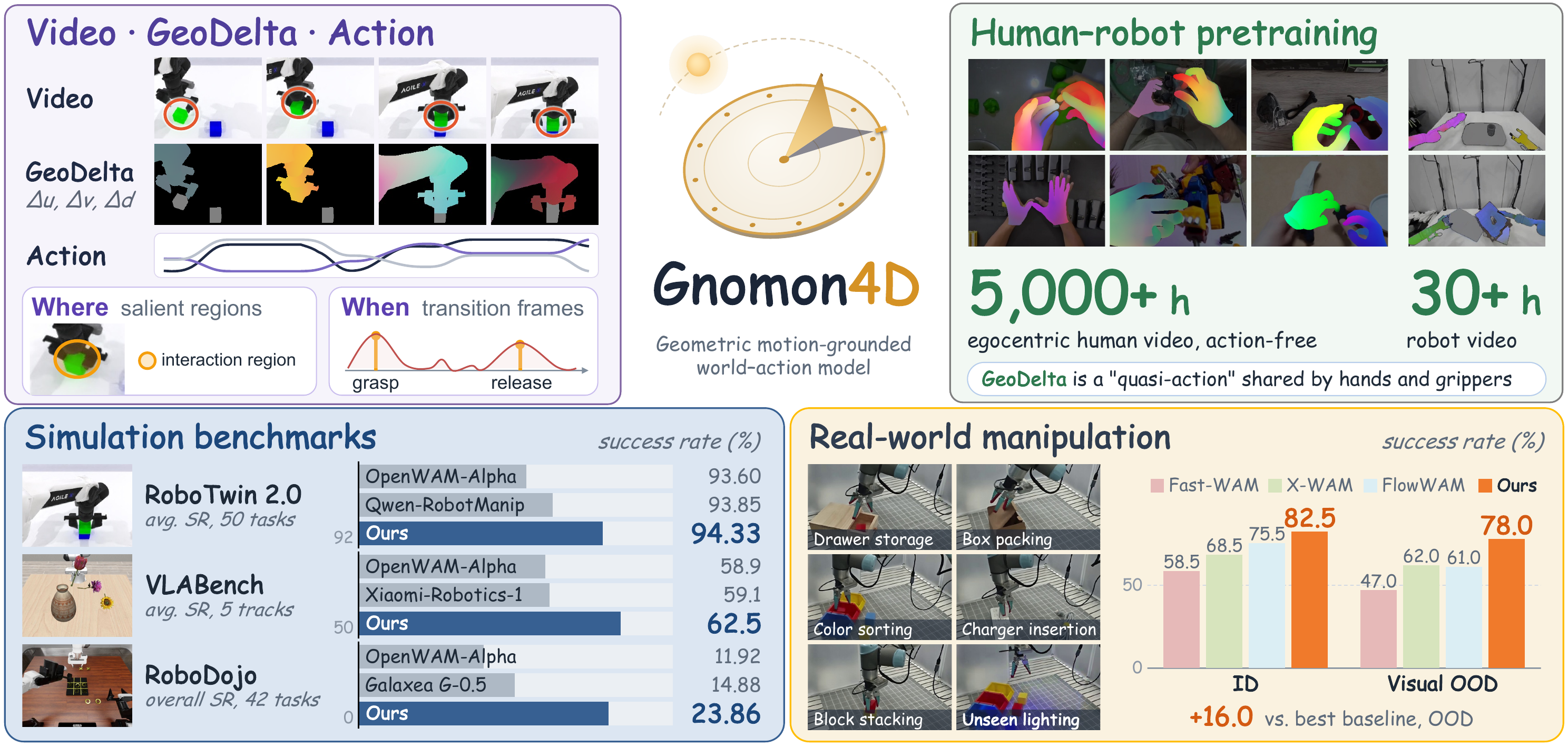}
  \captionof{figure}{\ours{} teaches the action where and when to focus: it jointly generates future video, actions, and \trk{}, the 4D motion $(\Delta u,\Delta v,\Delta d)$ of embodiment and object points, and reads \trk{} at interaction regions and transition frames (top left).
  As a human--robot quasi-action, \trk{} links pretraining on 5,000+ hours of egocentric video with mid-training on 30+ hours of robot data (top right).
  \ours{} leads in simulation (bottom left) and on a real robot, especially under visual shifts (bottom right).}
  \label{fig:teaser}
  \end{center}
}]

\maketitle

{\let\thefootnote\relax\footnotetext{\noindent $^{*}$~Work done during an internship at KlingAI Research.}}
{\let\thefootnote\relax\footnotetext{\noindent \dag~Project Leader.}}
{\let\thefootnote\relax\footnotetext{\noindent \ddag~Corresponding Author. Email: qyw@scut.edu.cn.}}

\begin{abstract}

World-action models (WAMs) predict scene changes to guide robot actions, yet their RGB futures entangle state changes with appearance and lack 3D geometry.
Recent WAMs add geometry as a training signal or a control interface, or build native 3D models, but decide what geometry to model rather than how the action should read it.
Intuitively, control-relevant cues lie at embodiment--object interactions and their transitions, yet probing existing WAMs reveals that their action attention underweights these interactions in favor of the rest of the embodiment.
To focus the action on these interactions, we present Gnomon4D, a WAM that jointly generates future video, robot actions, and GeoDelta, a video-aligned 4D geometric motion field recording the image-plane displacement and depth change of embodiment and object points.
With the embodiment and objects in one field, their relative motion, which marks an interaction, can be read off.
Guided by this motion and the task, Interaction-Salient Attention (InSAttn) selects where the action reads, and Transition-Anchored Aggregation (TransAgg) selects when, around grasp and release.
As GeoDelta encodes surface motion rather than embodiment-specific controls, it serves as a human--robot quasi-action that links action-free pretraining on EgoManip-5000H, which we curate, to robot mid-training on Robot-30H.
Gnomon4D leads on RoboTwin~2.0, VLABench, and RoboDojo, and outperforms real-robot baselines, especially under visual shifts.

\end{abstract}

\section{Introduction}
\label{sec:intro}

Robot manipulation advances along complementary routes: vision-language-action models (VLAs) map observations and instructions to actions~\cite{nvidia2025gr00tn1openfoundation,intelligence2026pi07steerablegeneralistrobotic}, world-action models (WAMs) jointly predict future observations and actions~\cite{shen2025videovlavideogeneratorsgeneralizable,zhu2025unifiedworldmodelscoupling,guo2026videopredictionpolicy2,wang2026openwamopenmodularexploration,miao2026internw0deltaworldactionmodel}, and agentic recursive self-improvement (RSI) refines skills or generates simulation-ready worlds~\cite{physicalrsi2026,physicalrsiteam2026awomosimdataengineagenticsimulationreadyworldgeneration}.
All three aim to generate actions that change object states to accomplish a task, and WAMs in particular explicitly predict these changes to guide the action.
Most WAMs represent this future as video~\cite{kim2026cosmospolicyfinetuningvideo,wang2026openwamopenmodularexploration,zhao2026rynnworld4d4dembodiedworld}, inheriting the visual priors of large video foundation models~\cite{wan2025wanopenadvancedlargescale}.
RGB futures, however, entangle task-relevant state changes with texture, lighting, and background, leading policies to overfit to appearance and generalize poorly to unseen visual conditions~\cite{yuan2026dreamwamrgbfutureprediction,lou2026maskworldmodelpredicting}.
As 2D projections of the scene, they further lack the 3D geometry that precise manipulation such as insertion and stacking requires~\cite{qu2025spatialvlaexploringspatialrepresentations}.

\begin{figure}[!t]
  \centering

  \includegraphics[width=\linewidth]{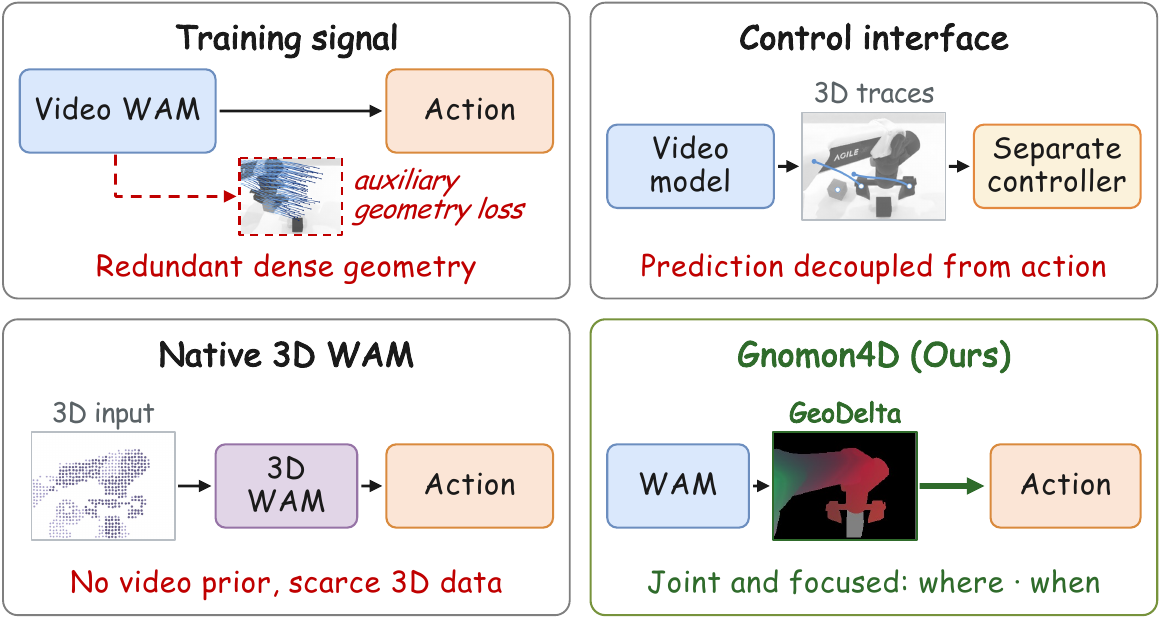}
  \caption{\textbf{Three routes to geometry in WAMs.} Each route leaves a limitation (red). \ours{} (bottom right) generates \trk{} jointly with video and actions and reads it where and when interactions occur, addressing the two limitations in the top row. Because \trk{} is embodiment-agnostic, it also enables human-video pretraining and avoids the scarce 3D data limiting native 3D WAMs.}
  \label{fig:paradigm}
\end{figure}

Recent work brings geometry into WAMs along three routes (Fig.~\ref{fig:paradigm}).
The first uses it as a \emph{training signal}: depth or 4D trajectory fields supervise the video features~\cite{guo2026unified4dworldaction,bardhan2026depthworld3dworldmodel,li2026wam4dfast4dworld,yang20264dwaminfusingspatiotemporalawareness,zhou2026gem4dgeometryenhancedvideoworld,zhang2026learning4dgeometricpriors}.
This supervision is dense over the whole scene, mostly irrelevant to the task, and leaves no explicit motion for the action to read: depth is frame-wise, and 4D fields are discarded at inference.
The second uses it as a \emph{control interface}: 3D traces, point dynamics, or object flow are predicted and passed to a separate controller~\cite{Lee_2026_CVPR_C2601,lee2026mu0scalable3dinteractiontrace,tong2026pointaction3dpointsuniversal,dharmarajan2025dream2flowbridgingvideogeneration}, which decouples world prediction from action generation.
The third route, \emph{native 3D WAMs}, builds on 3D or geometric backbones instead of video generators and predicts no future video~\cite{wang2026unitas3dnativeworldaction,park2026pointwam3dworldaction,han2026geometricactionmodelrobot,song2026roboticmanipulationvisiontogeometrymapping}, forgoing video priors and depending on far scarcer 3D data.

More fundamentally, these designs decide \emph{what} geometry to model but not \emph{how} the action should read it.
Intuitively, control-relevant cues concentrate in space at interaction-salient regions, where an object moves differently from the embodiment, and in time at transitions of this relation, such as grasp and release.
A plain Mixture-of-Transformers (MoT)~\cite{liang2025mixtureoftransformerssparsescalablearchitecture} WAM, however, leaves sparse, low-dimensional action supervision to find these cues among dense video and geometry tokens on its own, so the action expert tends to memorize embodiment-centric patterns instead~\cite{lin2026breakingvisionactionshortcutlatent}.
Probing the action attention of three released WAMs~\cite{yuan2026fastwamworldactionmodels,guo2026unified4dworldaction,wang2026openwamopenmodularexploration} on RoboTwin 2.0~\cite{chen2025robotwin20scalabledata} confirms this tendency (see the \textbf{supplementary material}): the rest of the embodiment draws up to 2.3$\times$ uniform attention, more than the interaction region.

To make these interactions explicit for the action, we introduce \textbf{\ours{}}, an MoT-based WAM that jointly generates future video, robot actions, and a video-aligned 4D geometric motion, \ie, \trk{} (Fig.~\ref{fig:teaser}).
For each foreground pixel on the embodiment and the manipulated object, \trk{} records the image-plane displacement and depth change $(\Delta u,\Delta v,\Delta d)$ of the same physical point over a short interval.
Because the embodiment and the objects share one motion field, their relative motion, the very cue that marks an interaction, can be read off directly.
Existing geometric futures cannot provide this cue: per-frame depth and point maps lack correspondence across frames, and 2D optical flow misses motion along the line of sight.
Defined on the video pixel grid, \trk{} also reuses the video VAE and stays aligned with the predicted frames and their generative priors.
A dedicated expert generates it jointly with video and actions and shares the denoising clock of the Action expert, so the action reads the predicted motion as it forms.

Yet explicit 4D geometric motion representation alone does not tell the action expert where and when to focus.
\ours{} therefore predicts an embodiment--object relative-motion field from \trk{} and, conditioned on the task, uses it to focus the action readout on interaction regions and moments.
Interaction-Salient Attention (InSAttn) selects where to read by weighting tokens with task-conditioned relative motion, and Transition-Anchored Aggregation (TransAgg) selects when by pooling frames around changes in this relation, such as grasp and release.
The resulting interaction tokens enter the Action expert residually, alongside its dense joint attention.

Finally, \trk{} is embodiment-agnostic: it describes foreground surface motion without encoding embodiment-specific joints, poses, or control parameters, and can therefore serve as a shared quasi-action across human hands and diverse robot embodiments.
Existing human-video pretraining pipelines instead treat estimated hand poses, such as MANO parameters, as actions~\cite{luo2026beingh07latentworldactionmodel,dyna2026dyna2}, learn latent actions from human videos~\cite{yang2026lawm3dlearning3dawarelatent}, or align human and robot data through embodiment-specific retargeting and robot rendering~\cite{ma2026humanscaleegocentrichumanvideo,miao2026internw0deltaworldactionmodel}.
By contrast, \trk{} is a video-aligned motion space of the embodiment and objects, shared directly by human and robot videos without mapping either source into the other's action space.
We curate \humanego{}, 5,000+ hours of egocentric manipulation video filtered from 37K hours of real-scene footage, and process Robot-30H, 30+ hours of real-robot demonstrations, constructing \trk{} supervision for both (see the \textbf{supplementary material}).
With \trk{} as the common target, pretraining on \humanego{} teaches the Video and \trk{} experts how an embodiment moves and interacts with objects without any action labels, so mid-training on Robot-30H can focus on grounding this motion space in robot controls before post-training on target tasks.

Our contributions are threefold:
\begin{itemize}
  \item \textbf{Joint world--motion--action generation.} \ours{}, a WAM that jointly generates future video, \trk{}, and actions, where \trk{} is a video-aligned 4D geometric motion field that makes embodiment--object interaction explicit to the action expert.
  \item \textbf{Relation-guided interaction readout.} InSAttn and TransAgg, which use the predicted embodiment--object relative motion and the task to focus the action expert on interaction regions and transition frames.
  \item \textbf{Embodiment-agnostic pretraining.} \trk{} as a human--robot quasi-action that links action-free pretraining on \humanego{} to robot mid-training on Robot-30H. \ours{} achieves state-of-the-art results on three simulation benchmarks and in real-robot experiments.
\end{itemize}

\section{Related Work}
\label{sec:related_work}

\paragraph{World-action models.}
WAMs use predicted futures for control, decoding actions from video features or jointly denoising video and actions~\cite{shen2025videovlavideogeneratorsgeneralizable,zhu2025unifiedworldmodelscoupling,ye2026worldactionmodelszeroshot,kim2026cosmospolicyfinetuningvideo,wang2026openwamopenmodularexploration,zhang2026nativevideoactionpretraininggeneralizable}, usually with an MoT backbone~\cite{liang2025mixtureoftransformerssparsescalablearchitecture}.
Fast-WAM and GigaWorld-Policy-0.5 train with future supervision but skip imagination at inference~\cite{yuan2026fastwamworldactionmodels,gigaworldteam2026gigaworldpolicy05fasterstrongerwam}, VPP2 distills a visual planner from event-level video pretraining~\cite{guo2026videopredictionpolicy2}, and InternW0-$\Delta$ learns Causal Imprint queries from 20,000+ hours of robot and human data~\cite{miao2026internw0deltaworldactionmodel}.
We keep imagination at inference, yet actions are ready halfway through sampling (Sec.~\ref{sec:joint}).

\paragraph{Geometry in world-action models.}
Geometry enters WAMs along three routes (Sec.~\ref{sec:intro}).
As a training signal, it supervises video features as an extra predicted target, such as RGB-D~\cite{guo2026unified4dworldaction,bardhan2026depthworld3dworldmodel}, pointmaps or geometric latents~\cite{Xu_2026_CVPR_C2605,hung2026modalityautoregressiveworldactionmodels}, and optical flow~\cite{zhao2026rynnworld4d4dembodiedworld,chen2026flowwamopticalflowunified}, or as depth, trajectory, and 4D priors distilled only during training~\cite{li2026wam4dfast4dworld,yang20264dwaminfusingspatiotemporalawareness,zhou2026gem4dgeometryenhancedvideoworld,miao2026internw0deltaworldactionmodel,wang2026track4actiondistillingworldcentric3d,Qian_2026_CVPR_C2611,zhang2026learning4dgeometricpriors}.
Frame-wise depth and pointmaps lack temporal correspondence, 2D flow omits line-of-sight motion, and distilled priors are discarded at inference.
Control-interface methods pass dynamic pointmaps, sparse 3D traces, or object flow to separate decoders or controllers~\cite{tong2026pointaction3dpointsuniversal,Lee_2026_CVPR_C2601,lee2026mu0scalable3dinteractiontrace,dharmarajan2025dream2flowbridgingvideogeneration}, decoupling prediction from action generation.
Native 3D methods model the world as 3D points, Gaussians, or 4D latents, or build on geometric backbones~\cite{Huang_2026_CVPR_C2602,xiong2026geomvlaunifyingscenemotion,chai2026gaf,li2026structured4dlatentpredictive,han2026geometricactionmodelrobot,song2026roboticmanipulationvisiontogeometrymapping,wang2026unitas3dnativeworldaction,park2026pointwam3dworldaction}, forgoing video priors and relying on scarcer 3D data.
\ours{} keeps the video prior and generates \trk{} jointly with video and actions for the action to read at inference.

\paragraph{Spatiotemporal readout for control.}
Beyond what the future encodes, other methods shape what the action reads from it.
Some compress the future into predictive states, joint embeddings, latent intentions or subgoals, learned queries, or an interaction-dynamics bottleneck~\cite{gu2026rightfutureactionlearning,miao2026internw0deltaworldactionmodel,jiang2026rethinkingrepresentationsworldactionmodeling,lin2026jepawamlearningvisionlanguageactionpolicies,li2026latentactionintentionenables,chen2026lawamlatentworldaction,yang2026lilawamlightweightlatentreasoning,Yang_2026_CVPR_C2613,Wang_2026_CVPR_C2603}.
Spatial selection draws on 3D grounding, affordances, interaction-aligned pruning, active attention, or end-effector distance~\cite{zhang2026groundedactionmodel3d,you2026affordancewamaffordanceawarejointworldaction,shao2026kamwmkinematicaffordancemaps,Liu_2026_CVPR_C2620,cheng2026trainingfreeinteractionalignedvisualtoken,Xiao_2026_CVPR_C2621,tian2026starryspatialtemporalactioncentricworld}.
Temporal designs organize the future by latent motion chains, subtask progress, or rolling denoising across replanning cycles~\cite{Yang_2026_CVPR_C2606,Liu_2026_CVPR_C2620,zhou2026rollingwamworldactionmodels}.
Interaction-aware objectives reweight world-model supervision toward dynamic regions~\cite{tang2026impactattentioninteractionmap,song2026kineworldactioninducedtransportfields} rather than guide the readout.
These cues are semantic, embodiment-centric, or learned, and methods selecting in both space and time use separate signals~\cite{Liu_2026_CVPR_C2620}.
InSAttn and TransAgg instead derive both from predicted embodiment--object relative motion.

\paragraph{Human-video pretraining.}
Most pipelines map human motion into robot action spaces by treating estimated hand poses as actions~\cite{luo2026beingh07latentworldactionmodel,dyna2026dyna2} or through retargeting and robot rendering~\cite{zheng2026egoscalescalingdexterousmanipulation,ma2026humanscaleegocentrichumanvideo,yuan2026qwenrobotmaniptechnicalreportalignment,miao2026internw0deltaworldactionmodel}.
Others learn action-free 3D-aware latent actions~\cite{yang2026lawm3dlearning3dawarelatent} or co-train WAMs with pixel, semantic, or 3D-flow world targets~\cite{li2026egowamworldactionmodels}.
Cross-embodiment methods transfer 4D affordances, contact trajectories, or tool--target motion from human demonstrations to robots~\cite{han2026a4acrossembodimenttransferactionoriented,azirar2026contactflowvideoactionconditioning,han2026bridgeactbridginghumandemonstrations}.
\trk{} is instead an explicit quasi-action that human and robot videos share without cross-embodiment mapping.

\section{Method}
\label{sec:method}

\begin{figure*}[!t]
  \centering
  \includegraphics[width=\textwidth]{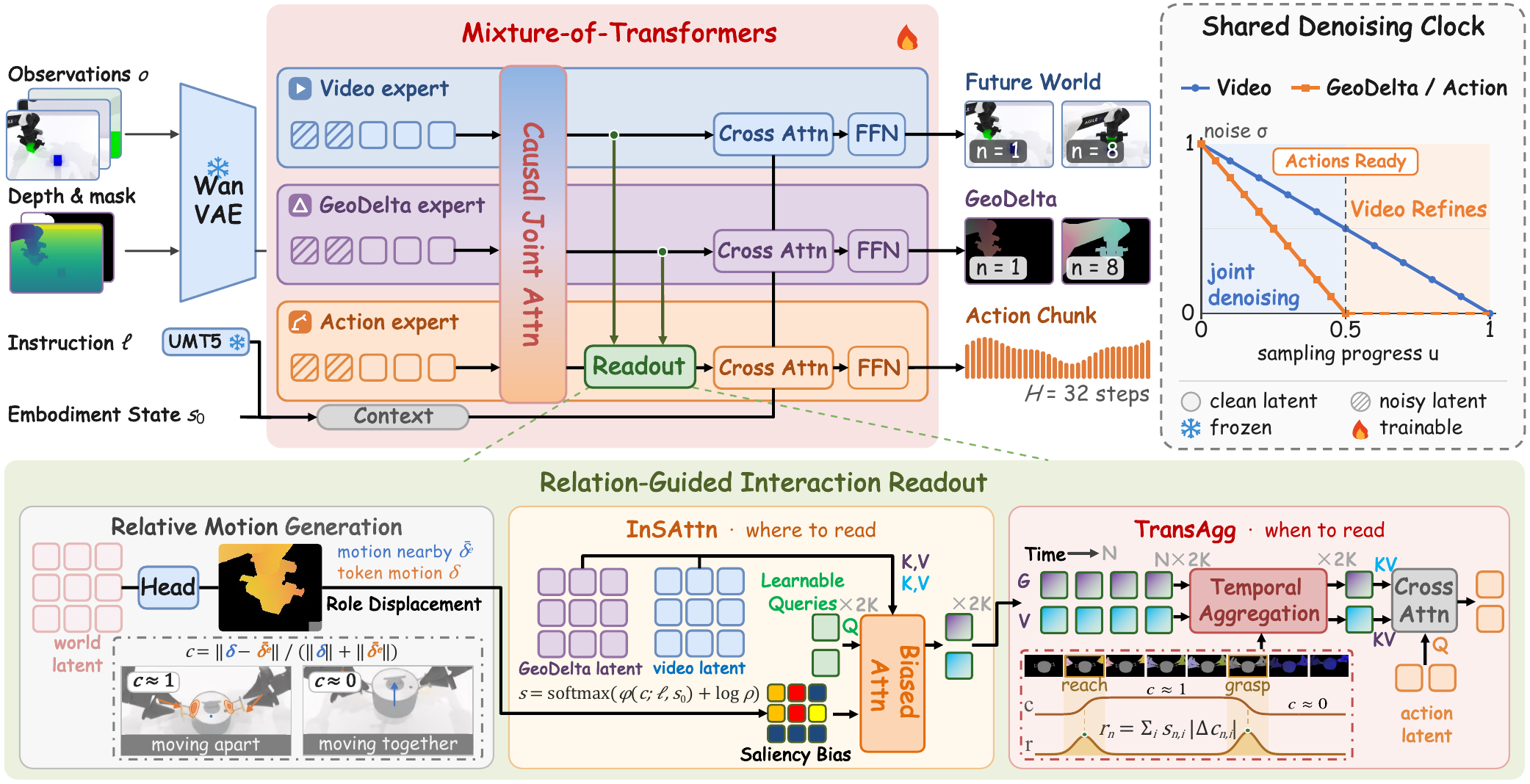}
  \caption{\textbf{Overview of \ours{}.}
  (a) The Video, \trk{}, and Action experts generate their future modalities through joint attention, and an interaction readout enters the Action expert residually.
  (b) \trk{} and Action share a denoising clock that finishes ahead of Video, so actions are ready before the video is refined.
  (c) InSAttn turns task-conditioned embodiment--object relative motion into saliency that selects interaction regions, and TransAgg aggregates the frames where this relation changes.
  Image tiles show RoboTwin examples.}
  \label{fig:pipeline}
\end{figure*}

\subsection{Preliminaries}
\label{sec:prelim}

WAMs built on video generators jointly denoise future video and actions~\cite{zhu2025unifiedworldmodelscoupling,guo2026unified4dworldaction,wang2026openwamopenmodularexploration}.
Each modality $m$ follows a flow-matching path~\cite{lipman2023flowmatchinggenerativemodeling} $\mathbf{x}^m_{\sigma_m}=(1-\sigma_m)\mathbf{x}^m+\sigma_m\boldsymbol{\epsilon}^m$ between a clean sample $\mathbf{x}^m$ (a video latent or an action chunk) and Gaussian noise $\boldsymbol{\epsilon}^m$, and the network $\mathbf{u}_\theta$ regresses the velocity given the context $\mathcal{C}$:
\begin{equation}
  \mathcal{L}^{m}_{\mathrm{FM}}
  =\mathbb{E}\,\big\|\mathbf{u}^{m}_\theta\big(\{\mathbf{x}^{m'}_{\sigma_{m'}}\}_{m'};\mathcal{C}\big)-(\boldsymbol{\epsilon}^m-\mathbf{x}^m)\big\|_2^2.
  \label{eq:fm}
\end{equation}
With modality-specific noise levels~\cite{zhu2025unifiedworldmodelscoupling,guo2026unified4dworldaction}, a modality at $\sigma_m=0$ serves as a clean condition and incurs no loss, so one network can act as a policy, an action-conditioned world model, or a joint generator.
The denoiser is typically a Mixture-of-Transformers (MoT)~\cite{liang2025mixtureoftransformerssparsescalablearchitecture}, whose modality experts have separate parameters and exchange information through joint attention at every layer.

\subsection{Overview}
\label{sec:overview}

At each replanning step, \ours{} receives multi-view observations $\mathbf{o}$, the proprioceptive state $\mathbf{s}_0$, and a language instruction $\ell$, denoted jointly as the context $\mathcal{C}$.
It predicts the next $H=32$ control steps in three modalities: an action chunk $\mathbf{A}=\{\mathbf{a}_\tau\}_{\tau=1}^{H}$, future video $\mathbf{V}=\{\mathbf{v}_n\}_{n=1}^{N}$, and a \trk{} sequence $\mathbf{G}=\{\mathbf{g}_n\}_{n=1}^{N}$.
To reduce computation, video and \trk{} are sampled every $\Delta=4$ control steps, giving $N=H/\Delta=8$ future frames.
\ours{} models the joint distribution $p_\theta(\mathbf{V},\mathbf{G},\mathbf{A}\mid\mathcal{C})$ with Video, \trk{}, and Action experts (Fig.~\ref{fig:pipeline}).

\subsection{\trk{}: 4D Geometric Motion}
\label{sec:track}

\trk{} describes the short-term 4D motion of foreground surfaces between adjacent sampled frames.
Let $\Pi_n$ map a 3D point to its pixel coordinates and depth in the camera of frame $n$.
For a foreground surface point that moves from $\mathbf{X}_n$ to $\mathbf{X}_{n+1}$, \trk{} stores at its source pixel $(u_n,v_n)$ the displacement
$\mathbf{g}_{n+1}(u_n,v_n)=\Pi_n(\mathbf{X}_{n+1})-\Pi_n(\mathbf{X}_n)=(\Delta u,\Delta v,\Delta d)$.

Measuring both positions in one camera makes the field invariant to camera motion.
Pixels without valid depth or correspondence are masked out.
The foreground consists of the embodiment and the manipulated objects, so action-driven embodiment motion and the resulting object motion share one pixel grid and time axis and can be compared directly.
Describing surface motion rather than joints or control signals, \trk{} also places human hands and robot embodiments in one supervision space (Sec.~\ref{sec:train}).

To reuse the frozen video VAE, we encode the three components $(\Delta u,\Delta v,\Delta d)$ of \trk{} as the three channels of an RGB video.
Each component $g_c$ is soft-thresholded to suppress label noise, normalized by a fixed scale $\omega_c$, and mapped to an 8-bit value $I_c$ by signed $\mu$-law companding:
\begin{equation}
  \begin{aligned}
    z_c&=\operatorname{clip}\!\left(g_c/\omega_c,-1,1\right),\\
    I_c&=127.5\left(1+\operatorname{sgn}(z_c)\frac{\log(1+\mu|z_c|)}{\log(1+\mu)}\right),
  \end{aligned}
  \label{eq:codec}
\end{equation}
where $\mu=31$, $\omega_u=\omega_v=W/6$ for image width $W$, and $\omega_d$ is set from the depth-change statistics of the training data.
The $\mu$-law assigns more intensity levels to small displacements and thus preserves the fine motion near contact.

\subsection{Joint World--Motion--Action Generation}
\label{sec:joint}

\paragraph{Architecture.}
\ours{} extends the Video--Action MoT (Sec.~\ref{sec:prelim}) with a \trk{} expert (Fig.~\ref{fig:pipeline}a): Video imagines the future appearance, and \trk{} supplies its geometry and motion for the Action expert to turn into execution.
Each expert is a stack of DiT blocks, and all three read the language instruction and proprioceptive state through cross-attention.
Both world experts denoise $N$ future latents, $\mathbf{V}$ and $\mathbf{G}$, conditioned on clean tokens of the current frame: the multi-view frame latent $\mathbf{V}_{\mathrm{obs}}$ for Video, and the encoded head-camera depth and foreground mask $\mathbf{G}_{\mathrm{cond}}$ for \trk{}.
Lacking condition tokens of its own, the Action expert observes the scene through $\mathbf{V}_{\mathrm{obs}}$ and $\mathbf{G}_{\mathrm{cond}}$ and denoises the $H$-step chunk $\mathbf{A}$.

\paragraph{Denoising clock and causal mask.}
For the three experts to reinforce one another, joint attention must expose each modality at the right time, which is set by the denoising clock and the attention mask.
Video needs many denoising steps for its fine detail, whereas low-dimensional actions need only a few, so a joint sampler should finish the actions first.
Sampling the noise level of each expert independently, as in standard MoT training, spreads training over all noise combinations, most of which inference never meets~\cite{guo2026unified4dworldaction}.
We instead tie the noise levels $\sigma_m$ of Eq.~\eqref{eq:fm} to one sampling progress $u$, which runs from $0$ at pure noise to $1$ at clean outputs.
\trk{} and actions are both sparse motion signals coupled through kinematics, so they share a clock that runs twice as fast as that of Video:
\begin{equation}
  \sigma_{\mathrm{V}}(u)=1-u,\quad
  \sigma_{\mathrm{G}}(u)=\sigma_{\mathrm{A}}(u)=\max(1-2u,0).
  \label{eq:clock}
\end{equation}
Sampling thus denoises all three modalities jointly in its first half.
Once $\sigma_{\mathrm{G}}=\sigma_{\mathrm{A}}=0$, \trk{} and actions serve as clean conditions while Video refines its details.
Training draws 40\% of samples from this clock and the rest from any $\sigma_{\mathrm{G}}=\sigma_{\mathrm{A}}\le\sigma_{\mathrm{V}}$, which also covers generating video from clean actions.
The attention mask follows the causal order of execution: future Video and \trk{} tokens attend only to actions up to their own time step, and clean tokens never attend to noisy ones (see the supplementary material).

\subsection{Relation-Guided Interaction Readout}
\label{sec:read}

Joint attention exposes all future tokens to the Action expert but does not tell it which ones matter for control.
As argued in Sec.~\ref{sec:intro}, control-relevant cues concentrate where an object moves differently from the nearby embodiment and when this relation changes, such as at grasp and release.
We therefore read the predicted future with $K=32$ learnable queries $\{\mathbf{q}_k\}_{k=1}^{K}$ per world modality.
Each query reads every frame with InSAttn focusing it on interaction regions, and TransAgg pools these readings around transitions.

\paragraph{Relative-motion field.}
Let $\mathbf{f}_{n,i}$ be the hidden feature of token $i$ in future frame $n$, taken from the Video or \trk{} expert.
Lightweight heads on \trk{} features predict, for both readout branches, a \emph{token motion} $\boldsymbol{\delta}_{n,i}\in\mathbb{R}^3$ and the probabilities $\rho^{\mathrm{e}}_{n,i}$ and $\rho^{\mathrm{o}}_{n,i}$ that the token belongs to the embodiment or to an object.
The local embodiment motion $\bar{\boldsymbol{\delta}}^{\mathrm{e}}_{n,i}$ is the average motion of embodiment tokens, such as the arm and gripper, around token $i$.
The relative-motion field $c$ measures each token's departure from it:
\begin{equation}
  c_{n,i}=\frac{\|\boldsymbol{\delta}_{n,i}-\bar{\boldsymbol{\delta}}^{\mathrm{e}}_{n,i}\|}{\|\boldsymbol{\delta}_{n,i}\|+\|\bar{\boldsymbol{\delta}}^{\mathrm{e}}_{n,i}\|+\varepsilon}\in[0,1].
  \label{eq:coupling}
\end{equation}
Thus $c$ is near $1$ where an object and the embodiment move relative to each other and near $0$ where they move together.
A grasp switches $c$ from the first case to the second and a release switches it back, so both show up as a large frame-to-frame change $|\Delta c_{n,i}|$, where $\Delta c_{n,i}=c_{n,i}-c_{n-1,i}$.

\begin{figure*}[t]
  \centering
  \includegraphics[width=0.88\textwidth]{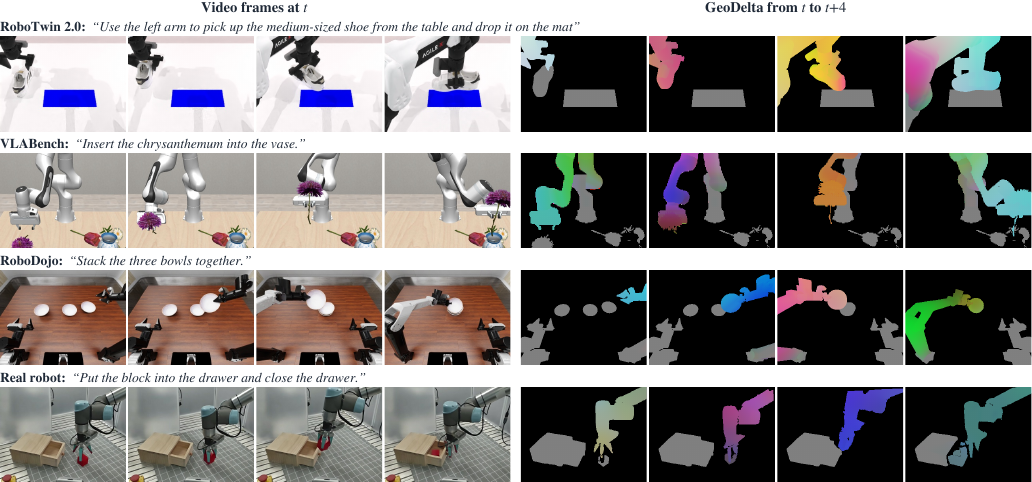}
  \caption{\trk{} supervision in the four domains. Each row shows four head-camera frames of one training demonstration with its instruction, and the \trk{} from each frame $t$ to $t{+}4$ encoded by Eq.~\eqref{eq:codec}. Neutral gray marks static embodiment or object pixels, and black marks background.}
  \label{fig:geodelta}
\end{figure*}

\paragraph{InSAttn: where to read.}
InSAttn turns $c$ into a saliency map over each frame.
A small MLP $\phi$ scores every token from its relative motion $c$ given the instruction $\ell$ and proprioception $\mathbf{s}_0$, and the log foreground probability suppresses the background:
\begin{equation}
  s_{n,i}=\operatorname*{softmax}_{i}\big(\phi(c_{n,i};\,\ell,\mathbf{s}_0)+\log(\rho^{\mathrm{e}}_{n,i}+\rho^{\mathrm{o}}_{n,i})\big).
  \label{eq:saliency}
\end{equation}
Because the score depends on motion rather than appearance, $s$ can single out the grasped object and the fingers around it instead of the most conspicuous pixels.
Each learnable query $\mathbf{q}_k$ then reads every frame $n$ separately:
\begin{equation}
  \mathbf{y}_{k,n}=\sum_i\operatorname*{softmax}_{i}\big(\mathbf{q}_k^{\top}\mathbf{f}_{n,i}+\log s_{n,i}\big)\,\mathbf{f}_{n,i}.
  \label{eq:frame_read}
\end{equation}
With key and value projections omitted for brevity, this is standard attention in which $\log s$ scales each token's attention weight by its saliency, giving a $K\times N$ grid $\mathbf{y}_{k,n}$.

\paragraph{TransAgg: when to read.}
TransAgg pools this grid over time, turning the $N$ per-frame vectors $\mathbf{y}_{k,1},\dots,\mathbf{y}_{k,N}$ of each query into one token $\mathbf{z}_k$.
It scores each frame by how much $c$ changes on salient tokens, $r_n=\sum_i s_{n,i}|\Delta c_{n,i}|$, which peaks at transitions such as grasp and release.
Each query then aggregates its vectors as in Eq.~\eqref{eq:frame_read}, with $\log r_n$ in place of $\log s_{n,i}$:
\begin{equation}
  \mathbf{z}_k=\sum_n\operatorname*{softmax}_{n}\big(\mathbf{q}_k^{\top}\mathbf{y}_{k,n}+\log r_n\big)\,\mathbf{y}_{k,n},
  \label{eq:readout}
\end{equation}
so transition frames dominate each token (pseudocode in the supplementary material).

\paragraph{Injection.}
The $2K$ tokens $\mathbf{Z}=\{\mathbf{z}^{\mathrm{V}}_k,\mathbf{z}^{\mathrm{G}}_k\}$ of the two modalities enter the Action layers through a residual cross-attention alongside the dense joint attention.
During training, an auxiliary head predicts clean actions from $\mathbf{Z}$ alone while the Action expert's instruction and proprioception are randomly masked, forcing $\mathbf{Z}$ to carry action-relevant evidence.

\subsection{Training and Inference}
\label{sec:train}

\paragraph{Training.}
Because human hands and robot embodiments share the \trk{} space, the world experts can learn how an embodiment interacts with objects from human video alone.
We therefore pretrain the Video and \trk{} experts on \humanego{} without action labels.
We then add the Action expert, mid-train the full model on Robot-30H to ground this motion space in robot controls, and post-train it on each simulated and real-world benchmark.

\paragraph{Inference.}
All modalities start from noise and follow the clock of Eq.~\eqref{eq:clock} over 20 solver steps, with noise levels shifted as in Wan2.2~\cite{wan2025wanopenadvancedlargescale}.
\trk{} and actions become clean after step 10, when the controller executes a prefix of the action chunk and replans from the next observation.
The remaining steps refine Video only when a visual prediction is needed.

\section{Experiments}
\label{sec:exp}

We evaluate whether explicit 4D geometric motion improves closed-loop control, where the gains emerge, and which components of \ours{} produce them.
The study covers three simulation benchmarks, five real-robot tasks under in-distribution (ID) and out-of-distribution (OOD) conditions, and ablations of the future representation, interaction readout, and pretraining curriculum.

\subsection{Experimental Setup}
\label{sec:exp_setup}

\paragraph{Baselines.}
In simulation, we compare with representative VLA and WAM models (Tables~\ref{tab:robotwin}--\ref{tab:robodojo}).
On RoboTwin 2.0, WAMs are further grouped by how they represent and read out predicted futures: RGB futures, 3D geometry or flow (depth, 4D geometric priors, or 2D optical flow), and structured readout that selects action-relevant information.
The real-robot baselines are Fast-WAM~\cite{yuan2026fastwamworldactionmodels}, X-WAM~\cite{guo2026unified4dworldaction}, and FlowWAM~\cite{chen2026flowwamopticalflowunified}, whose futures are RGB, RGB-D, and RGB with 2D flow, respectively.

\paragraph{\trk{} supervision.}
\trk{} labels require per-pixel depth, embodiment and object masks, and cross-frame correspondences, which public demonstrations do not provide in full (details per benchmark in the supplementary material).
We replay RoboTwin 2.0 demonstrations from their seeds in SAPIEN and regenerate VLABench demonstrations in MuJoCo with the official scene sampler and expert skills.
Moving each back-projected pixel with the recorded rigid motion of its robot link or object then yields exact \trk{} for both the embodiment and the objects.
RoboDojo releases only RGB videos, camera parameters, and robot states, without the object poses or scene seeds needed to reproduce its scenes.
Forward kinematics of the released joint states still gives exact embodiment \trk{}, whereas object \trk{} is estimated with DA3~\cite{lin2025depth3recoveringvisual} depth, SAM~3.1~\cite{carion2026sam3segmentconcepts} masks, and RAFT~\cite{teed2020raftrecurrentallpairsfield} flow, with DA3 depth rescaled per episode to match the rendered embodiment depth.
On the real robot, \trk{} is computed from measured RGB-D depth, SAM~3.1 masks, and RAFT flow (Fig.~\ref{fig:geodelta}).
At inference, \ours{} takes the current RGB image, depth, and foreground mask as input (Sec.~\ref{sec:joint}): simulator-rendered conditions on RoboTwin 2.0 and VLABench, robot-only renders from joint states on RoboDojo, and RGB-D depth with a SAM~3.1 mask on the real robot.

\paragraph{Implementation details.}
We initialize the Video expert from Wan2.2-TI2V-5B~\cite{wan2025wanopenadvancedlargescale} (30 layers, width 3072) and derive the \trk{} and Action experts (width 1024) from it by width interpolation, so that all three share its attention space.
We train with FSDP hybrid sharding in bfloat16.

\subsection{Simulation Experiments}
\label{sec:exp_sim}

\begin{table*}[t]
  \centering
  \begin{minipage}[t]{\columnwidth}
    \centering
\caption{Success rate (\%) on RoboTwin 2.0-Full over 50 bimanual tasks. Clean and Rand.\ denote the clean and domain-randomized settings. WAMs are grouped by their future-state representation and readout mechanism.}
\label{tab:robotwin}
\footnotesize
\setlength{\tabcolsep}{16.82pt}
\setlength{\aboverulesep}{0.55ex}
\setlength{\belowrulesep}{0.80ex}
\renewcommand{\arraystretch}{1.10}
\begin{tabular}{@{}lccc@{}}
\toprule
\textbf{Method} & \textbf{Clean} & \textbf{Rand.} & \textbf{Avg.} \\
\midrule
\rowcolor{gray!10}[0pt][0pt]
\multicolumn{4}{@{}l@{}}{\rule[-0.75ex]{0pt}{3.3ex}\textbf{VLA}} \\
$\pi_{0.5}$~\cite{black2025pi05visionlanguageactionmodelopenworld} & 82.74 & 76.76 & 79.75 \\
Galaxea G-0.5~\cite{liu2026g05autoregressivestreamrobot} & 93.72 & 92.84 & 93.28 \\
Qwen-RobotManip~\cite{yuan2026qwenrobotmaniptechnicalreportalignment} & 93.70 & 94.00 & 93.85 \\
\midrule
\rowcolor{gray!10}[0pt][0pt]
\multicolumn{4}{@{}l@{}}{\rule[-0.75ex]{0pt}{3.3ex}\textbf{WAM: RGB Futures}} \\
Motus~\cite{Bi_2026_CVPR_C2604} & 88.66 & 87.02 & 87.84 \\
Fast-WAM~\cite{yuan2026fastwamworldactionmodels} & 91.88 & 91.78 & 91.83 \\
LingBot-VA 2.0~\cite{zhang2026nativevideoactionpretraininggeneralizable} & 93.80 & 93.40 & 93.60 \\
OpenWAM-Alpha~\cite{wang2026openwamopenmodularexploration} & 93.74 & 93.46 & 93.60 \\
\midrule
\rowcolor{gray!10}[0pt][0pt]
\multicolumn{4}{@{}l@{}}{\rule[-0.75ex]{0pt}{3.3ex}\textbf{WAM: 3D Geometry / Flow}} \\
X-WAM~\cite{guo2026unified4dworldaction} & 89.76 & 90.68 & 90.22 \\
WAM4D~\cite{li2026wam4dfast4dworld} & 93.82 & 89.86 & 91.84 \\
FlowWAM~\cite{chen2026flowwamopticalflowunified} & 92.94 & 92.14 & 92.54 \\
MECo-WAM~\cite{zhang2026learning4dgeometricpriors} & 93.26 & 91.98 & 92.62 \\
\midrule
\rowcolor{gray!10}[0pt][0pt]
\multicolumn{4}{@{}l@{}}{\rule[-0.75ex]{0pt}{3.3ex}\textbf{WAM: Structured Readout}} \\
ARPS~\cite{gu2026rightfutureactionlearning} & 91.32 & 90.10 & 90.71 \\
STARRY~\cite{tian2026starryspatialtemporalactioncentricworld} & 93.82 & 93.30 & 93.56 \\
\midrule
\rowcolor{cvprblue!10}
\multicolumn{1}{@{}>{\columncolor{cvprblue!10}[0pt][\tabcolsep]}l}{\textbf{\ours{}}} & \textbf{94.56} & \textbf{94.10} & \multicolumn{1}{>{\columncolor{cvprblue!10}[\tabcolsep][0pt]}c@{}}{\textbf{94.33}} \\
\bottomrule
\end{tabular}

  \end{minipage}\hfill
  \begin{minipage}[t]{\columnwidth}
    \centering
\caption{Success rate (\%) across VLABench tracks.}
\label{tab:vlabench}
\footnotesize
\setlength{\tabcolsep}{3.8pt}
\setlength{\aboverulesep}{0.40ex}
\setlength{\belowrulesep}{0.65ex}
\renewcommand{\arraystretch}{1.13}
\begin{tabular}{@{}llcccccc@{}}
\toprule
\multicolumn{2}{@{}l}{\textbf{Method}} & \textbf{ID} & \textbf{Cat} & \textbf{CS} & \textbf{Ins} & \textbf{Tex} & \textbf{Avg.} \\
\midrule
\multirow{2}{*}{\textbf{VLA}} & $\pi_{0.5}$~\cite{black2025pi05visionlanguageactionmodelopenworld} & 65.4 & 38.2 & 43.9 & 48.2 & 44.9 & 48.1  \\
 & Xiaomi-Robotics-1~\cite{xiaomiroboticsteam2026xiaomirobotics1scalingvisionlanguageactionmodels} & 75.6 & \textbf{53.0} & 48.4 & 55.8 & 62.6 & 59.1  \\
\midrule
\textbf{WAM} & OpenWAM-Alpha~\cite{wang2026openwamopenmodularexploration} & 83.4 & 38.1 & 58.0 & 53.8 & 61.4 & 58.9 \\
\midrule
\rowcolor{cvprblue!10}
\multicolumn{2}{@{}>{\columncolor{cvprblue!10}[0pt][\tabcolsep]}l}{\textbf{\ours{}}} & \textbf{84.0} & 39.1 & \textbf{61.7} & \textbf{59.1} & \textbf{68.8} & \multicolumn{1}{>{\columncolor{cvprblue!10}[\tabcolsep][0pt]}c@{}}{\textbf{62.5}} \\
\bottomrule
\end{tabular}

    \vspace{9.9pt}
    \centering
\caption{Success rate (\%) and Score on RoboDojo.}
\label{tab:robodojo}
\footnotesize
\setlength{\tabcolsep}{1.96pt}
\setlength{\aboverulesep}{0.40ex}
\setlength{\belowrulesep}{0.65ex}
\renewcommand{\arraystretch}{1.13}
\begin{tabular}{@{}lccccccc@{}}
\toprule
\multirow{2}{*}{\textbf{Method}} & \multicolumn{5}{c}{\textbf{Dimension SR}} & \multicolumn{2}{c}{\textbf{Overall}} \\
\cmidrule(lr){2-6}\cmidrule(l){7-8}
 & \textbf{Gen.} & \textbf{Prec.} & \textbf{Long} & \textbf{Mem.} & \textbf{Open} & \textbf{SR} & \textbf{Score} \\
\midrule
\rowcolor{gray!10}[0pt][0pt]
\multicolumn{8}{@{}l@{}}{\rule[-0.6ex]{0pt}{3ex}\textbf{VLA}} \\
InternVLA-A1.5~\cite{ma2026internvlaa15unifyingunderstandinglatent} & 6.83 & 10.17 & 13.75 & 3.56 & 1.42 & 7.14 & 11.15 \\
$\pi_{0.5}$~\cite{black2025pi05visionlanguageactionmodelopenworld} & 8.17 & 5.50 & 14.67 & 4.67 & 1.67 & 6.93 & 11.44 \\
Xiaomi-Robotics-1~\cite{xiaomiroboticsteam2026xiaomirobotics1scalingvisionlanguageactionmodels} & 17.00 & 18.83 & 23.67 & 6.56 & 3.58 & 13.93 & 20.07 \\
Galaxea G-0.5~\cite{liu2026g05autoregressivestreamrobot} & 12.83 & 20.42 & \textbf{32.25} & 7.33 & 1.58 & 14.88 & 20.23 \\
\midrule
\rowcolor{gray!10}[0pt][0pt]
\multicolumn{8}{@{}l@{}}{\rule[-0.6ex]{0pt}{3ex}\textbf{WAM}} \\
Fast-WAM~\cite{yuan2026fastwamworldactionmodels} & 1.11 & 0.00 & 5.17 & 3.44 & 0.42 & 2.03 & 3.48 \\
GigaWorld-Policy-0~\cite{ni2026gigaworldpolicyefficientactioncentered} & 2.89 & 1.83 & 8.92 & 2.22 & 0.50 & 3.27 & 6.20 \\
X-WAM~\cite{guo2026unified4dworldaction} & 3.33 & 1.83 & 9.08 & 4.67 & 0.25 & 3.83 & 7.69 \\
OpenWAM-Alpha~\cite{wang2026openwamopenmodularexploration} & 14.83 & 9.25 & 25.33 & 9.11 & 1.08 & 11.92 & 17.18 \\
\midrule
\rowcolor{cvprblue!10}
\multicolumn{1}{@{}>{\columncolor{cvprblue!10}[0pt][\tabcolsep]}l}{\textbf{\ours{}}} & \textbf{23.28} & \textbf{28.24} & 24.44 & \textbf{34.34} & \textbf{9.02} & \textbf{23.86} & \multicolumn{1}{>{\columncolor{cvprblue!10}[\tabcolsep][0pt]}c@{}}{\textbf{30.96}} \\
\bottomrule
\end{tabular}

  \end{minipage}
\end{table*}

\paragraph{Setup.}
We train one multitask policy per benchmark and report success rate (SR) throughout.
RoboTwin 2.0~\cite{chen2025robotwin20scalabledata} evaluates bimanual manipulation on 50 tasks in clean and randomized scenes.
VLABench~\cite{zhang2024vlabenchlargescalebenchmarklanguageconditioned} tests language understanding and commonsense reasoning with 10 task families on five tracks: ID, unseen categories (Cat), commonsense variations (CS), semantic instructions (Ins), and unseen textures (Tex).
RoboDojo~\cite{chen2026robodojounifiedsimandrealbenchmark} spans 42 bimanual tasks over Generalization, Precision, Long-Horizon, Memory, and Open-Instruction, and macro-averages SR and a partial-credit Score over them.

\paragraph{Results.}
\ours{} achieves the best average result on all three benchmarks.
On the near-saturated RoboTwin 2.0 (Table~\ref{tab:robotwin}), it ranks first in both settings and surpasses MECo-WAM, the strongest geometry- or flow-based WAM, by 1.71 points.
MECo-WAM uses 4D targets only as a training signal, whereas \ours{} generates \trk{} at inference for the action to read.
Against STARRY, which biases action attention by distance to the predicted end effector, reading embodiment--object relative motion adds 0.77 points.

The gains are pronounced under appearance shifts and in precise positioning.
On VLABench (Table~\ref{tab:vlabench}), the margin over OpenWAM-Alpha grows from 0.6 points on ID to 7.4 on unseen textures, as \trk{} encodes motion rather than appearance.
RoboDojo (Table~\ref{tab:robodojo}) shows the second effect.
\ours{} doubles the SR of OpenWAM-Alpha (23.86\% vs.\ 11.92\%), and on the Precision dimension it reaches 28.24\% against at most 20.42\% for the baselines, as these tasks demand exact positions in both the image plane and depth, which \trk{} encodes jointly.

\subsection{Real-Robot Experiments}
\label{sec:exp_real}

\paragraph{Setup.}
We use a UR5e arm with a parallel gripper, a fixed third-person RGB-D camera, and a wrist camera.
On this platform, we evaluate five tasks covering routine manipulation (Drawer Storage, Box Packing, and Block Stacking), long-horizon manipulation (Color Sorting), and precision manipulation (Charger Insertion).
All methods are trained on the same 50 demonstrations per task (250 in total).
Each method is then evaluated in 80 trials per task (400 in total) from initial configurations shared by all methods.
Half of the trials follow the training conditions (ID), and the other half are OOD, with 10 trials for each of four shifts: unseen lighting, tablecloth, or object appearance, and added distractors (details in the supplementary material).

\begin{figure}[t]
  \centering
  \includegraphics[width=\linewidth]{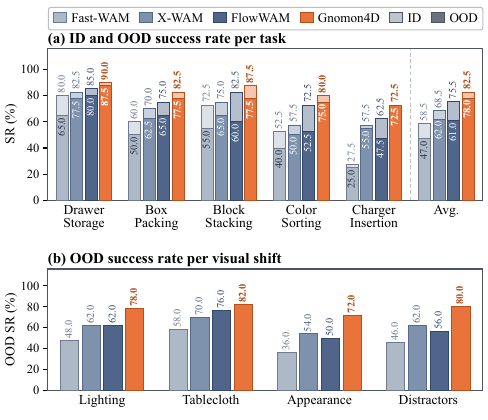}
  \caption{Real-robot success rate. (a) SR per task and its average over the five tasks (Avg.), with each method's OOD bar (dark) overlaid on its ID bar (light). (b) OOD SR under each visual shift, aggregated over 50 trials across the five tasks. Success counts are reported in the supplementary material.}
  \label{fig:real_results}
\end{figure}

\paragraph{Results.}
\ours{} ranks first on every task under both conditions, reaching 82.5\% ID and 78.0\% OOD SR (Fig.~\ref{fig:real_results}a).
The baselines trade accuracy for robustness: FlowWAM leads them in ID (75.5\%) but loses 14.5 points under visual shifts, whereas X-WAM loses only 6.5 but trails in ID.
Encoding image-plane and depth motion jointly, \trk{} gives \ours{} both: it loses only 4.5 points, and its lead over the runner-up grows from 7.0 points in ID to 16.0 in OOD.

Among the visual shifts, the margin is largest under unseen object appearance and added distractors (18.0 points each, Fig.~\ref{fig:real_results}b), both of which leave \trk{} unchanged: it does not depend on texture, and a static distractor contributes no motion.
Across tasks, Color Sorting and Charger Insertion gain the most under OOD conditions (22.5 and 17.5 points), as Color Sorting compounds errors over three interactions and Charger Insertion demands exact depth.

\subsection{Ablation Study}
\label{sec:exp_abl}

Ablations use 10 interaction-intensive RoboTwin 2.0 tasks, with every variant post-trained for 20k steps.
The three groups in Table~\ref{tab:ablation} are cumulative: starting from the Wan2.2-5B initialization without pretraining, (a) varies the future representation under dense readout, (b) varies the readout on full \trk{}, and (c) adds the pretraining stages.

\begin{table}[t]
    \centering
    \caption{Ablations on 10 interaction-intensive RoboTwin 2.0 tasks.}
    \label{tab:ablation}
    \footnotesize
    \setlength{\tabcolsep}{5pt}
    \renewcommand{\arraystretch}{1.05}
    \begin{tabular}{@{}lccc@{}}
    \toprule
    \textbf{Variant} & \textbf{Clean} & \textbf{Rand.} & \textbf{Avg.} \\
    \midrule
    \rowcolor{gray!10}[0pt][0pt]
    \multicolumn{4}{@{}l@{}}{\rule[-0.6ex]{0pt}{3ex}\textbf{(a) Future Representation (Dense Readout)}} \\
    RGB & 81.0 & 79.0 & 80.0 \\
    RGB + Depth & 81.5 & 80.3 & 80.9 \\
    RGB + 2D Flow & 83.1 & 81.9 & 82.5 \\
    RGB + \trk{} (Body) & 81.5 & 79.8 & 80.7 \\
    RGB + \trk{} (Full) & 83.8 & 82.9 & 83.4 \\
    \midrule
    \rowcolor{gray!10}[0pt][0pt]
    \multicolumn{4}{@{}l@{}}{\rule[-0.6ex]{0pt}{3ex}\textbf{(b) Readout Structure}} \\
    Generic Queries & 84.2 & 82.9 & 83.6 \\
    InSAttn Only & 85.7 & 84.8 & 85.3 \\
    InSAttn + TransAgg & 86.6 & 85.7 & 86.2 \\
    \midrule
    \rowcolor{gray!10}[0pt][0pt]
    \multicolumn{4}{@{}l@{}}{\rule[-0.6ex]{0pt}{3ex}\textbf{(c) Pretraining Curriculum}} \\
    Wan2.2-5B Init. (No Pretraining) & 86.6 & 85.7 & 86.2 \\
    + \humanego{} & 88.4 & 88.2 & 88.3 \\
    \rowcolor{cvprblue!10}
    \multicolumn{1}{@{}>{\columncolor{cvprblue!10}[0pt][\tabcolsep]}l}{\textbf{+ Robot-30H (\ours{})}} & \textbf{91.3} & \textbf{90.6} & \multicolumn{1}{>{\columncolor{cvprblue!10}[\tabcolsep][0pt]}c@{}}{\textbf{91.0}} \\
    \bottomrule
    \end{tabular}
    \end{table}

\textbf{Object motion drives the gain of \trk{}.}
With embodiment motion alone, \trk{} barely improves over RGB (+0.7), while adding object motion brings another 2.7 points (Table~\ref{tab:ablation}a).
\trk{} thus helps by capturing how objects respond to the embodiment rather than by re-encoding the action.
Full \trk{} also exceeds depth and 2D flow by 2.5 and 0.9 points, and its margin over flow grows from 0.7 points in Clean to 1.0 in Rand., where varying table heights make the same image-plane displacement correspond to different 3D motion.
\textbf{Guiding where and when to read matters more than query capacity.}
Generic queries merely match dense readout (83.6\% vs.\ 83.4\%), whereas InSAttn and TransAgg add 1.7 and 0.9 points (Table~\ref{tab:ablation}b).
\textbf{Human video transfers through \trk{}.}
Pretraining the world experts on \humanego{} improves SR by 2.1 points (Table~\ref{tab:ablation}c), and mid-training on Robot-30H adds another 2.7.

\section{Conclusion}
\label{sec:conclusion}

We presented \ours{}, a WAM that jointly generates future video, robot actions, and \trk{}, a video-aligned 4D geometric motion field of embodiment and object points.
Because the embodiment and objects share this field, their relative motion marks each interaction, and InSAttn and TransAgg use it with the task to select where and when the action reads.
As a human--robot quasi-action, \trk{} also links action-free pretraining on \humanego{} to robot mid-training on Robot-30H.
\ours{} leads on three simulation benchmarks and a real robot, with the clearest gains in precise manipulation such as insertion and under visual shifts.
Ablations show that \trk{} helps mainly through object motion and that the guided readout outperforms generic queries, suggesting that a predicted future serves control through both what it encodes and how the action reads it.

\paragraph{Limitations and future work.}

Without simulator state, robot \trk{} is estimated from measured or monocular depth, masks, and optical flow, whose errors on thin, reflective, or transparent surfaces mainly perturb $\Delta d$.
Human-video \trk{} covers only the hands and is rendered from fitted MANO meshes, which compensate camera rotation but not translation and omit the depth change of the hand root.
High-quality 4D annotations of real-world manipulation therefore remain urgently needed.
One scalable source is agentic recursive self-improvement, which could rebuild real scenes as simulation-ready worlds where \trk{} is exact or refine the estimation pipeline with embodied feedback.

\section*{Acknowledgements}
This work was supported by KlingAI, the National Natural Science Foundation of China (NSFC) under Grant Nos. 62672187 and 62125107, and the Guangdong Basic and Applied Basic Research Foundation under Grant No. 2025B1515120037.

{
    \small
    \bibliographystyle{ieeenat_fullname}
    \bibliography{gnomon4d_ref}
}

\clearpage
\makeatletter
\global\Hy@pagecounter=\numexpr\value{page}-1\relax
\makeatother
\hypersetup{hypertexnames=false}
\setcounter{page}{1}
\setcounter{section}{0}
\renewcommand{\thesection}{\Alph{section}}
\maketitlesupplementary
\flushbottom
\setlength{\textfloatsep}{12pt plus 2pt minus 2pt}
\setlength{\dbltextfloatsep}{12pt plus 2pt minus 2pt}
\setlength{\floatsep}{10pt plus 2pt minus 2pt}
\setlength{\dblfloatsep}{10pt plus 2pt minus 2pt}
\renewcommand{\topfraction}{0.9}
\renewcommand{\dbltopfraction}{0.9}
\renewcommand{\textfraction}{0.08}
\renewcommand{\floatpagefraction}{0.8}
\renewcommand{\dblfloatpagefraction}{0.75}
\makeatletter
\setlength{\@fptop}{0pt}
\setlength{\@dblfptop}{0pt}
\setlength{\@fpsep}{\floatsep}
\setlength{\@fpbot}{0pt plus 1fil}
\setlength{\@dblfpsep}{\dblfloatsep}
\setlength{\@dblfpbot}{0pt plus 1fil}
\makeatother
\providecommand{\cmark}{{\usefont{U}{pzd}{m}{n}\symbol{51}}}
\providecommand{\xmark}{{\usefont{U}{pzd}{m}{n}\symbol{55}}}
\providecommand{\ourscol}[1]{\multicolumn{1}{>{\columncolor{cvprblue!10}}c}{#1}}
\providecommand{\ourslast}[1]{\multicolumn{1}{>{\columncolor{cvprblue!10}[\tabcolsep][0pt]}c@{}}{#1}}

This supplement first examines where the action reads the predicted future (Sec.~\ref{sec:supp_probe}) and how existing WAMs differ in what they predict and how they read it (Sec.~\ref{sec:supp_compare}).
Section~\ref{sec:supp_method} then details the joint attention mask and interaction readout of \ours{}, and Sec.~\ref{sec:supp_datasets} describes the human and robot data used for pretraining and mid-training.
The last two sections give data construction and experimental protocols for simulation (Sec.~\ref{sec:supp_sim}) and the real robot (Sec.~\ref{sec:supp_real_setup}), including the real-robot success counts.

\section{Attention Probes}
\label{sec:supp_probe}

We examine whether the action readout of a WAM concentrates on embodiment--object interactions.
The study compares \ours{} with three released RoboTwin 2.0 policies, Fast-WAM, X-WAM, and OpenWAM-Alpha, without updating their weights.

\subsection{Probe Protocol}
\label{sec:supp_probe_protocol}

We use 50 clean-condition event windows, one per RoboTwin 2.0 task, with matched observations, instructions, and seeds.
The baselines are probed through their action-to-video attention.
For \ours{}, we record the interaction-token route of the benchmark checkpoint, whose 20-round schedule completes actions after round 10.
Its action-to-token, temporal, and spatial attention weights are composed, and the object-side and embodiment-side maps are then averaged.
For all models, attention is averaged over action queries, heads, active denoising steps, and readout layers, and then normalized within the head-camera domain.

Attention is measured in the future latent block containing the event.
Fast-WAM, whose action reads only the observed frame, is measured on that frame instead.
Simulator embodiment masks and motion-associated object components define two reference regions.
The interaction region consists of the motion-associated object and the embodiment within 3\,cm of contact, and the rest of the embodiment forms the second region.
Motion requires a surface displacement of 4\,mm over two steps and at least 40 moving pixels, and static regions follow their overlap with the nearest moving frame.
Contact uses 3\,cm proximity, and event selection also includes motion onset, lift-off (1.5\,cm), and settling, merging gaps of up to eight steps.

Reference masks are preprocessed to match each model's native input: a $384{\times}320$ three-view layout for Fast-WAM, OpenWAM-Alpha, and \ours{}, and three $256{\times}320$ views with a $0.95$ center crop for X-WAM.
Each $8{\times}10$ future grid pools four stride-four reference masks.
Attention enrichment divides the attention mass on a region by the region's area fraction, so a value of $1$ corresponds to uniform attention.
Bootstrap intervals resample tasks 5{,}000 times.

\subsection{Results and Visualization}
\label{sec:supp_probe_spatial}
\label{sec:supp_probe_comparison}

\begin{figure}[!t]
  \centering
  \includegraphics[width=\linewidth]{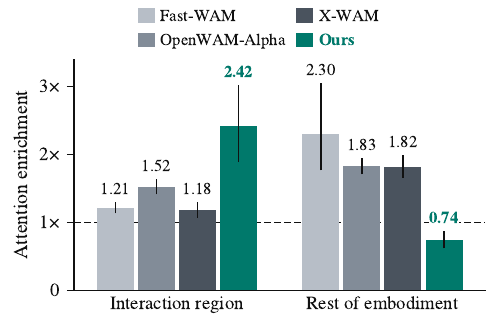}
  \caption{\textbf{Where does action attention go?} Attention enrichment on the interaction region, formed by the manipulated object and the embodiment within 3\,cm of contact, and on the rest of the embodiment, over 50 event windows, one per RoboTwin 2.0 task. Enrichment is the attention mass on a region divided by its area fraction, and the dashed line marks uniform attention (1$\times$). The baselines favor the embodiment as a whole, whereas \ours{} concentrates on the interaction region. Error bars are 95\% bootstrap intervals over tasks.}
  \label{fig:probe}
\end{figure}

\begin{figure*}[!t]
  \centering
  \includegraphics[width=\textwidth]{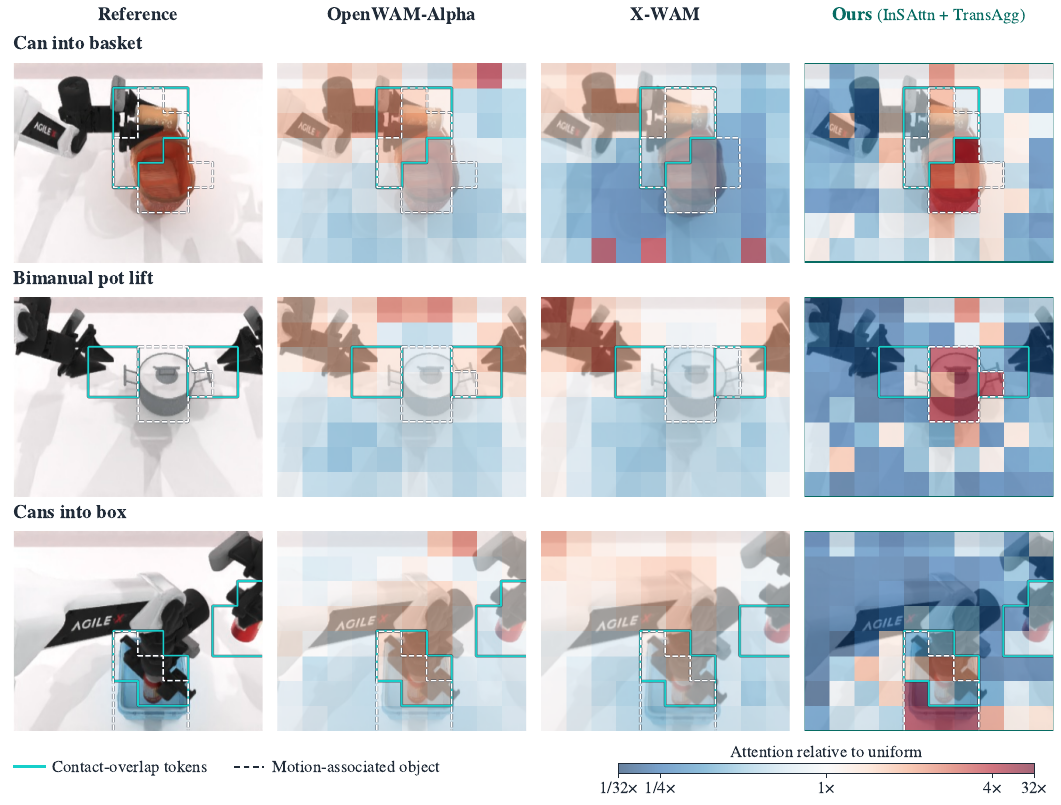}
  \caption{\textbf{Action-conditioned spatial readout.} Three interactions selected from the 50-window probe, whose population statistics are given in Sec.~\ref{sec:supp_probe_spatial}. All models are shown on the future latent block containing the event, and Fast-WAM is omitted because its action reads only the observed frame. For \ours{}, the maps show the interaction rollout of the benchmark checkpoint. Reference regions are pooled within the displayed block. Cyan contours mark contact-overlap tokens, and dashed white contours mark motion-associated objects. Colors show attention relative to uniform within each frame on the native $8{\times}10$ token grid, using one piecewise log scale from $\tfrac{1}{32}\times$ to $32\times$ in which $\tfrac{1}{4}\times$--$4\times$ spans 80\% of the color range.}
  \label{fig:probe_heatmap}
\end{figure*}

The baselines favor the embodiment as a whole (Fig.~\ref{fig:probe}): its remaining part receives 1.82--2.30$\times$ uniform attention, more than the 1.18--1.52$\times$ on the interaction region.
\ours{} reverses this allocation, raising the interaction region to 2.42$\times$ [1.89, 3.03] and lowering the rest of the embodiment to 0.74$\times$.
Attention to the embodiment near contact stays at 1.51$\times$, within the 1.27--1.77$\times$ range of the baselines.
The embodiment bias of the baselines motivates InSAttn, whose region selection in \ours{} moves the readout from the arm body to the interacting pair.

Figure~\ref{fig:probe_heatmap} shows three selected interactions under the same protocol.
On the 46 windows with visible objects, mean object enrichment is 2.53$\times$ for \ours{}, versus 1.48$\times$ for OpenWAM-Alpha and 1.16$\times$ for X-WAM, with paired gains of 1.05 [0.43, 1.73] and 1.37 [0.76, 2.05] (95\% bootstrap intervals).
\ours{} exceeds both baselines on 25 of the 46 windows, and its maps retain scattered off-object peaks.

\section{Design Comparison with Existing WAMs}
\label{sec:supp_compare}

\begin{table*}[t]
\centering
\caption{Design comparison with the WAM baselines of Table~1 of the main paper. \emph{Future}: prediction targets beyond RGB video. \emph{Online}: the action reads a predicted future at inference; otherwise the future is a training signal only. \emph{3D motion}: the target encodes cross-frame 3D displacement; EE, end effector only. \emph{Emb./Obj.}: embodiment and object motion are represented separately. \emph{Ego. video}: pretraining on egocentric human video. \emph{Params}: inference-time parameters, summed from the components each paper reports; n/r, not reported. Avg.\ is the RoboTwin 2.0 SR of Table~1 of the main paper.}
\label{tab:supp_design}
\footnotesize
\setlength{\tabcolsep}{3.9pt}
\renewcommand{\arraystretch}{1.08}
\begin{tabular}{@{}llccccllc@{}}
\toprule
\textbf{Method} & \textbf{Future} & \textbf{Online} & \textbf{3D motion} & \textbf{Emb./Obj.} & \textbf{Ego. video} & \textbf{Action readout} & \textbf{Params} & \textbf{Avg.} \\
\midrule
\rowcolor{gray!10}[0pt][0pt]
\multicolumn{9}{@{}l@{}}{\rule[-0.6ex]{0pt}{3ex}\textbf{WAM: RGB Futures}} \\
Motus~\cite{Bi_2026_CVPR_C2604} & Flow latent action & \cmark & \xmark & \xmark & \cmark & Joint attention & 8B & 87.84 \\
Fast-WAM~\cite{yuan2026fastwamworldactionmodels} & -- & \xmark & \xmark & \xmark & \xmark & Current frame only & 6B & 91.83 \\
LingBot-VA 2.0~\cite{zhang2026nativevideoactionpretraininggeneralizable} & Latent action & \cmark & \xmark & \xmark & \cmark & Joint attention & 15.3B (MoE) & 93.60 \\
OpenWAM-Alpha~\cite{wang2026openwamopenmodularexploration} & -- & \cmark & \xmark & \xmark & \cmark & Joint attention & 6B & 93.60 \\
\midrule
\rowcolor{gray!10}[0pt][0pt]
\multicolumn{9}{@{}l@{}}{\rule[-0.6ex]{0pt}{3ex}\textbf{WAM: 3D Geometry / Flow}} \\
X-WAM~\cite{guo2026unified4dworldaction} & Per-frame depth & \cmark & \xmark & \xmark & \xmark & Joint attention & n/r & 90.22 \\
WAM4D~\cite{li2026wam4dfast4dworld} & Per-frame depth & \xmark & \xmark & \xmark & n/r & History frames only & 5.7B & 91.84 \\
FlowWAM~\cite{chen2026flowwamopticalflowunified} & Robot 2D flow & \cmark & \xmark & \xmark & \cmark & Cross-attention & 5.8B & 92.54 \\
MECo-WAM~\cite{zhang2026learning4dgeometricpriors} & 4D geometry tokens & \xmark & \cmark & \xmark & \xmark & Current frame only & 6B & 92.62 \\
\midrule
\rowcolor{gray!10}[0pt][0pt]
\multicolumn{9}{@{}l@{}}{\rule[-0.6ex]{0pt}{3ex}\textbf{WAM: Structured Readout}} \\
ARPS~\cite{gu2026rightfutureactionlearning} & V-JEPA2 features & \xmark & \xmark & \xmark & n/r & Predictive-state queries & 6B & 90.71 \\
STARRY~\cite{tian2026starryspatialtemporalactioncentricworld} & Depth, EE pose & \cmark & EE & EE & \cmark & EE-distance attention bias & n/r & 93.56 \\
\midrule
\rowcolor{cvprblue!10}
\multicolumn{1}{@{}>{\columncolor{cvprblue!10}[0pt][\tabcolsep]}l}{\textbf{\ours{}}} & \trk{} $(\Delta u,\Delta v,\Delta d)$ & \cmark & \cmark & \cmark & \cmark & InSAttn + TransAgg & 6.76B & \multicolumn{1}{>{\columncolor{cvprblue!10}[\tabcolsep][0pt]}c@{}}{\textbf{94.33}} \\
\bottomrule
\end{tabular}
\end{table*}

The attention probes motivate examining both what a predicted future contains and how the action expert reads it.
Table~\ref{tab:supp_design} compares the WAM baselines evaluated in the main paper along three axes: future representation, access at inference, and action readout.

\paragraph{Representation.}
No baseline gives the action online access to the 3D motion of both the embodiment and the objects.
Futures that carry geometry or motion encode it frame by frame (X-WAM, WAM4D), restrict 3D motion to the end effector (STARRY), keep motion in the image plane (Motus, FlowWAM), or use 4D targets only as a training signal (MECo-WAM).
\trk{} is generated together with the action and records image displacement and depth change on a common grid for the embodiment and the objects.

\paragraph{Readout.}
Five baselines read the predicted future through dense joint attention or cross-attention, and STARRY biases action attention by distance to the predicted end effector.
Fast-WAM, WAM4D, MECo-WAM, and ARPS generate no future at inference.
Among them, ARPS reads current video-expert features through predictive-state queries trained against future V-JEPA2 targets.
InSAttn and TransAgg instead select regions and frames from the predicted embodiment--object relative motion, which neither the end-effector bias nor the predictive-state queries represent.
Nor do the gains come from scale: among models with reported sizes, \ours{} has 6.76B parameters, within the 5.7--8B of the baselines except LingBot-VA 2.0 (15.3B, mixture of experts).

\section{Model Implementation}
\label{sec:supp_method}

The joint attention mask determines which predictions each expert can access, and the interaction readout selects spatial and temporal evidence for the Action expert.

\begin{algorithm}[!b]
  \caption{Relation-guided interaction readout.}
  \label{alg:readout}
  \small
  \begin{algorithmic}[1]
    \Require future features $\mathbf{f}^m_{n,i}$, queries $\{\mathbf{q}^m_k\}_{k=1}^{K}$, and grid alignment $\mathcal{A}_m$, for $m\in\{\mathrm{V},\mathrm{G}\}$; instruction $\ell$ and proprioception $\mathbf{s}_0$
    \Ensure interaction tokens $\mathbf{Z}=\{\mathbf{z}^{\mathrm{V}}_k,\mathbf{z}^{\mathrm{G}}_k\}_{k=1}^{K}$
    \Statex \textit{Estimate geometric cues shared by both modalities.}
    \State $\boldsymbol{\delta}_{n,i},\rho^{\mathrm{e}}_{n,i},\rho^{\mathrm{o}}_{n,i}\gets\mathrm{heads}(\mathbf{f}^{\mathrm{G}}_{n,i})$
    \State $\bar{\boldsymbol{\delta}}^{\mathrm{e}}_{n,i}\gets\sum_{j}w_{ij}\rho^{\mathrm{e}}_{n,j}\boldsymbol{\delta}_{n,j}\big/\big(\sum_{j}w_{ij}\rho^{\mathrm{e}}_{n,j}+\varepsilon\big)$
    \State $c_{n,i}\gets\dfrac{\|\boldsymbol{\delta}_{n,i}-\bar{\boldsymbol{\delta}}^{\mathrm{e}}_{n,i}\|}{\|\boldsymbol{\delta}_{n,i}\|+\|\bar{\boldsymbol{\delta}}^{\mathrm{e}}_{n,i}\|+\varepsilon}$
    \State $\Delta c_{1,i}\gets0$ \Comment{No preceding future frame.}
    \State $\Delta c_{n,i}\gets c_{n,i}-c_{n-1,i}$ for $n=2,\ldots,N$
    \State $h_{n,i}\gets\phi(c_{n,i};\ell,\mathbf{s}_0)$
    \State $s_{n,i}\gets\operatorname{softmax}_i\big(h_{n,i}+\log(\rho^{\mathrm{e}}_{n,i}+\rho^{\mathrm{o}}_{n,i}+\varepsilon)\big)$
    \State $r_n\gets\sum_i s_{n,i}|\Delta c_{n,i}|$
    \Statex \textit{InSAttn: read salient regions within each frame.}
    \For{$m\in\{\mathrm{V},\mathrm{G}\}$}
      \State $\widetilde{s}^{m}\gets\mathcal{A}_m(s)$ \Comment{Match the feature grid.}
      \For{$k=1,\ldots,K$, $n=1,\ldots,N$}
        \State $\mathbf{y}^m_{k,n}\gets\operatorname{Attn}\big(\mathbf{q}^m_k,\{\mathbf{f}^m_{n,i}\}_i;\log(\widetilde{s}^{m}_{n,i}+\varepsilon)\big)$
      \EndFor
    \EndFor
    \Statex \textit{TransAgg: pool frames around relation changes.}
    \For{$m\in\{\mathrm{V},\mathrm{G}\}$, $k=1,\dots,K$}
      \State $\mathbf{z}^m_k\gets\operatorname{Attn}\big(\mathbf{q}^m_k,\{\mathbf{y}^m_{k,n}\}_n;\log(r_n+\varepsilon)\big)$
    \EndFor
    \State \Return $\{\mathbf{z}^{\mathrm{V}}_k,\mathbf{z}^{\mathrm{G}}_k\}_{k=1}^{K}$
  \end{algorithmic}
\end{algorithm}

\subsection{Attention Visibility}
\label{sec:supp_attn}

\begin{figure*}[!t]
  \centering
  \includegraphics[width=0.9\textwidth]{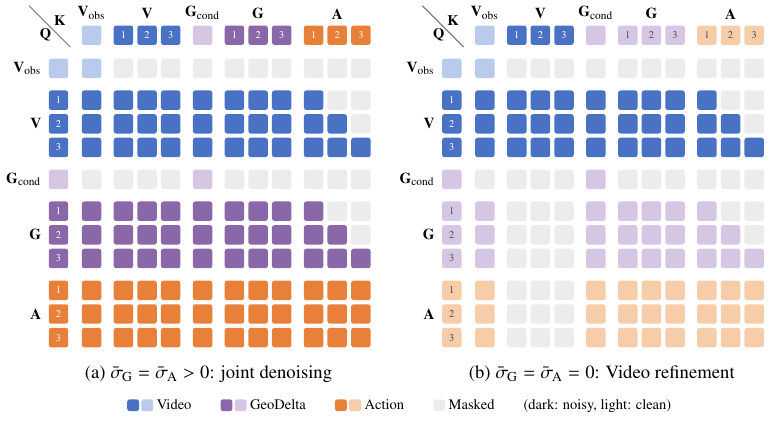}
  \caption{Attention visibility of the joint softmax. Rows are queries and columns are keys. Colored cells are visible and gray cells are masked, with dark and light shades marking noisy and clean blocks and numbers indexing time steps. (a) While \trk{} and actions are noisy, future Video and \trk{} tokens read all world tokens and the actions up to the corresponding control time, action tokens read the full sequence, and current-observation condition tokens read only themselves. (b) Future Video reads the clean \trk{} and Action blocks, which do not read noisy future Video.}
  \label{fig:attn_visibility}
\end{figure*}

Each layer applies one masked softmax over the three experts (Fig.~\ref{fig:attn_visibility}).
Its five token groups are clean current Video $\mathbf{V}_{\mathrm{obs}}$, future Video $\mathbf{V}$, clean \trk{} conditions $\mathbf{G}_{\mathrm{cond}}$, future \trk{} $\mathbf{G}$, and actions $\mathbf{A}$.
Current-observation tokens attend only within their own modality, so noisy predictions cannot alter them.
To learn the visual and geometric responses, future Video and \trk{} tokens read all world tokens and the actions up to the corresponding control time, whereas actions read themselves, the clean conditions, and all future world tokens.
A zero-noise block becomes a condition and stops attending to other modalities' noisy blocks.
During the second half of sampling, future Video therefore continues to read the clean \trk{} and actions, while these clean blocks no longer read noisy future Video (Fig.~\ref{fig:attn_visibility}b).
The same visibility rule applies when training with clean \trk{} and actions paired with noisy Video.

\subsection{Interaction Readout}
\label{sec:supp_readout}

\begin{figure*}[!tb]
  \centering
  \includegraphics[width=\textwidth]{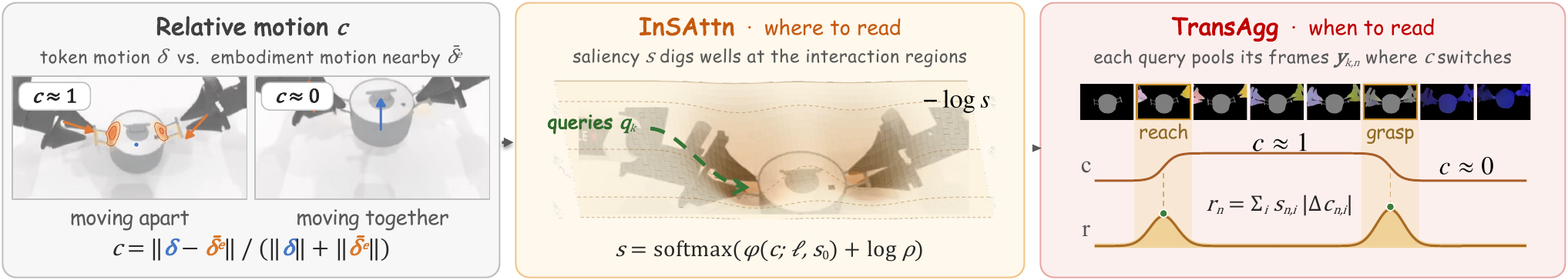}
  \caption{\textbf{Intuition behind relation-guided interaction readout.}
  Left: relative motion decreases when the embodiment and object move together.
  Middle: InSAttn converts this relation into task-conditioned spatial saliency.
  Right: TransAgg emphasizes frames where the relation changes.
  Here $\rho=\rho^{\mathrm{e}}+\rho^{\mathrm{o}}$ denotes foreground probability.}
  \label{fig:readout_intuition}
\end{figure*}

Figure~\ref{fig:readout_intuition} illustrates the readout on a reach-and-grasp interaction.
Relative motion increases as the gripper approaches a stationary object and decreases once the two move together.
Conditioned on the task, InSAttn maps this relation to saliency over the embodiment and the object.
The wells in the middle panel depict $-\log s$, so deeper wells mark a stronger attention bias.
TransAgg then pools frame features by the saliency-weighted change of the relation, concentrating evidence around interaction transitions.
Algorithm~\ref{alg:readout} reads the valid tokens of $N$ future frames at one readout layer and denoising step.

The motion and role heads act on \trk{} features and are supervised by an L1 loss on mean token displacement and a cross-entropy loss on role labels.
All geometric quantities are computed for every future frame $n$ and token $i$, with weights $w_{ij}$ that form a fixed $5{\times}5$ Gaussian neighborhood around $i$ and vanish outside it.
The MLP $\phi$ is FiLM-conditioned on $\ell$ and $\mathbf{s}_0$.

Each modality has $K=32$ queries, and both modalities share the saliency $s$ and frame scores $r$ computed on the \trk{} grid.
The operator $\mathcal{A}_m$ bilinearly resamples $s$ onto the feature grid of modality $m$, keeping the weights nonnegative and the frame index unchanged, and $\mathcal{A}_{\mathrm{G}}$ is the identity.
With projected keys and values, attention takes the form
\begin{equation*}
  \begin{aligned}
  &\operatorname{Attn}(\mathbf{q},\{\mathbf{x}_j\};b_j)=\sum_j\alpha_j W_V\mathbf{x}_j,\\
  &\alpha_j=\operatorname{softmax}_j\!\left(
  \frac{(W_Q\mathbf{q})^\top W_K\mathbf{x}_j}{\sqrt{d}}+b_j
  \right),
  \end{aligned}
\end{equation*}
where the projections are specific to each modality and attention stage and $d$ is the key dimension.
Both branches output tokens in a common readout space.
A small $\varepsilon>0$ is added before logarithms and in motion denominators, so when all $r_n$ vanish, the temporal bias is constant and pooling reduces to content-based attention.
The resulting $2K$ tokens enter the Action expert through residual cross-attention.

\section{Pretraining and Mid-Training Data}
\label{sec:supp_datasets}

The curriculum uses \humanego{} for action-free pretraining and Robot-30H for mid-training with actions.

\subsection{\humanego{} for Pretraining}
\label{sec:supp_data}

\begin{figure*}[!tbp]
  \centering
  \includegraphics[width=0.90\textwidth]{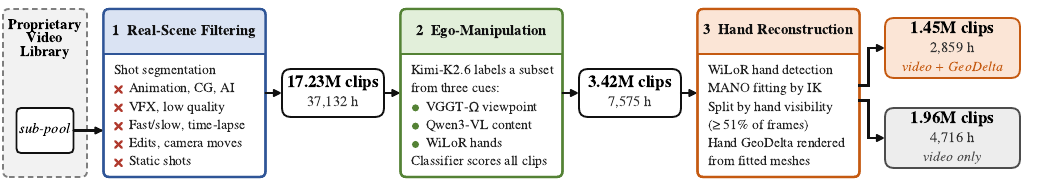}
  \caption{Curation of \humanego{}. A sub-pool of the library passes real-scene filtering and an egocentric-manipulation classifier, and hand reconstruction splits the result into a hand-visible group with \trk{} and a video-only group. Rounded boxes give the clips and hours that each stage outputs.}
  \label{fig:data_pipeline}
\smallskip
  \centering
  \includegraphics[width=0.90\textwidth]{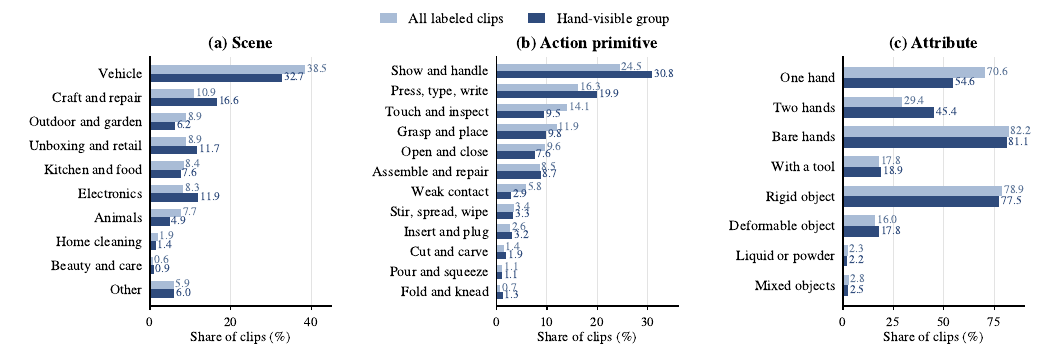}
  \caption{Composition of \humanego{}. Shares of Kimi-K2.6 scene, action, and attribute labels among labeled clips of the full set and of the hand-visible group.}
  \label{fig:data_composition}
\smallskip
  \centering
  \includegraphics[width=0.90\textwidth]{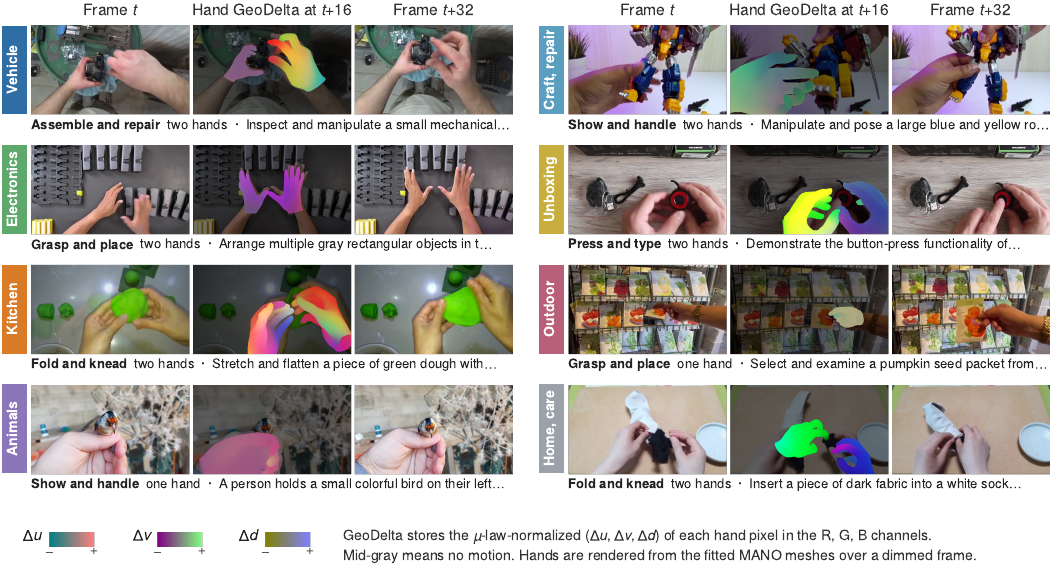}
  \caption{Training windows of \humanego{} across scenes. Each example shows frames $t$ and $t{+}32$ and, between them, the RGB-encoded hand \trk{} at $t{+}16$, labeled with the action primitive and the caption intent of its clip.}
  \label{fig:data_diversity}
\end{figure*}

\humanego{} is mined from a sub-pool of an industrial-scale proprietary video library, in which egocentric manipulation accounts for only about 5\% of the footage.
Three stages refine the sub-pool (Fig.~\ref{fig:data_pipeline}), and Table~\ref{tab:data_stats} reports the clips and hours that each stage retains.

\begin{table}[t]
\centering
\caption{Output of each curation stage of \humanego{}.}
\label{tab:data_stats}
\footnotesize
\setlength{\tabcolsep}{5pt}
\renewcommand{\arraystretch}{1.12}
\begin{tabular}{@{}lrrl@{}}
\toprule
\textbf{Stage} & \textbf{Clips} & \textbf{Hours} & \textbf{Supervises} \\
\midrule
1\enspace Real-scene filtering & 17.23M & 37,132 & -- \\
2\enspace Ego-manipulation classifier & 3.42M & 7,575 & -- \\
3\enspace Hand reconstruction & & & \\
\rowcolor{cvprblue!10}
\multicolumn{1}{@{}>{\columncolor{cvprblue!10}[0pt][\tabcolsep]}l}{\quad Hands visible} & 1.45M & 2,859 & \multicolumn{1}{>{\columncolor{cvprblue!10}[\tabcolsep][0pt]}l@{}}{Video, \trk{}} \\
\rowcolor{gray!10}
\multicolumn{1}{@{}>{\columncolor{gray!10}[0pt][\tabcolsep]}l}{\quad Hands rarely visible} & 1.96M & 4,716 & \multicolumn{1}{>{\columncolor{gray!10}[\tabcolsep][0pt]}l@{}}{Video} \\
\bottomrule
\end{tabular}
\end{table}

\paragraph{Real-scene filtering.}
Production video operators split each video into single-shot clips and keep only real-time recordings of real scenes.
A clip is discarded if it is animated, CG, or AI-generated, contains visual effects, or has low visual quality.
Static shots and clips whose timing or viewpoint has been altered by fast or slow motion, time-lapse, editing, or stylized camera moves are discarded as well.
The 17.23M surviving clips (37,132 hours) enter the next stage.

\paragraph{Egocentric manipulation classifier.}
This stage selects egocentric clips in which the camera wearer manipulates objects.
Running a large VLM on all 17.23M clips is prohibitively expensive, so a VLM labels only the training data of a lightweight classifier.
Kimi-K2.6~\cite{kimiteam2026kimik26} labels a subset of clips from three cues: camera trajectories and 3D geometry from VGGT-$\Omega$~\cite{wang2026vggtomega} reveal how the viewpoint moves, Qwen3-VL~\cite{bai2025qwen3vltechnicalreport} captions and embeddings describe the content, and WiLoR~\cite{potamias2025wilorendtoend3dhand} detections indicate whether hands are present.
A Qwen3-VL-Embedding-2B~\cite{li2026qwen3vlembedding} classifier trained on these labels scores every clip, and the 3.42M accepted clips (7,575 hours) form the egocentric manipulation set.

\paragraph{Hand reconstruction.}
This stage splits the egocentric manipulation set by whether the hands remain in view.
WiLoR detects the hands in every frame, and we fit a MANO~\cite{romero2017embodiedhands} mesh to each detection by inverse kinematics.
Clips whose hands appear in at least 51\% of the frames form the hand-visible group of 1.45M clips (2,859 hours), which supervises both the Video and \trk{} experts.
Their hand \trk{} is rendered from the fitted meshes during data loading, after the camera rotation between frames is compensated with the VGGT-$\Omega$ extrinsics.
The remaining 1.96M clips (4,716 hours) supervise only the Video expert.
Every clip also carries a Qwen3-VL caption with a global intent and time-stamped events, which prompts each training window.

\paragraph{Composition.}
We annotate a subset of high-confidence clips with Kimi-K2.6 scene and action labels and report their distribution in Fig.~\ref{fig:data_composition}.
Vehicles are the most common scene (38.5\%), and showing, pressing, inspecting, and grasping account for two thirds of the actions.
The hand-visible group contains more bimanual clips (45.4\% vs.\ 29.4\% overall), and Fig.~\ref{fig:data_diversity} shows examples from eight scene groups.

\subsection{Robot-30H for Mid-Training}
\label{sec:supp_robot}

\begin{figure*}[!t]
  \centering
  \includegraphics[width=\textwidth]{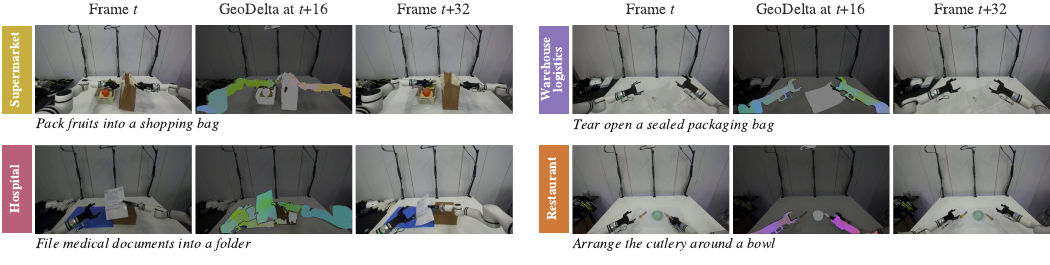}
  \caption{Robot-30H training windows from different scenes. Each example shows frames $t$ and $t{+}32$ and, between them, the RGB-encoded \trk{} of both arms and the manipulated objects at $t{+}16$, with parts at rest shown in mid-gray.}
  \label{fig:robot_examples}
\end{figure*}

\paragraph{Robot platform and tasks.}
We process Robot-30H from publicly released real-robot demonstrations~\cite{zhou2026imitatorgamebenchmarkingrobot}, recorded on a dual-arm Realman platform teleoperated in VR at 30\,Hz.
Each arm has seven joints and a parallel gripper, and a fixed ZED2i camera provides the head view that our model observes.
The demonstrations cover 53 tasks in household, supermarket, restaurant, warehouse logistics, hospital, and laboratory scenes.
Each task is recorded in several scene variants that change the object layout, appearance, or category, so the same skill appears in visually diverse settings.

\begin{figure}[t]
  \centering
  \includegraphics[width=\linewidth]{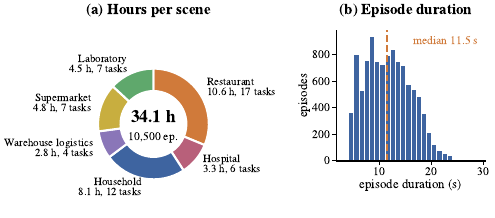}
  \caption{Robot-30H overview. (a) Hours of the 10,500 episodes by scene. (b) Distribution of episode durations.}
  \label{fig:robot_overview}
\end{figure}

\paragraph{Data scale and split.}
The source release contains 11,220 episodes and 37.0 hours of demonstrations.
We use 10,500 of these episodes (34.1 hours), which cover all 53 tasks (Fig.~\ref{fig:robot_overview}) with a median episode length of 11.5\,s.
The last two episodes of every task variant are held out for validation, leaving 10,080 episodes (32.8 hours) for training.

\paragraph{Training targets.}
Every episode is converted to the format of the post-training benchmarks, with the head view resized to $512{\times}288$ (Fig.~\ref{fig:robot_examples}).
The recordings provide depth only for the three auxiliary RealSense cameras and no camera calibration, so DA3~\cite{lin2025depth3recoveringvisual} estimates the metric depth and intrinsics of the ZED2i head view.
Released instance masks label the manipulated objects, and SAM~3.1~\cite{carion2026sam3segmentconcepts} segments the robot arms.
Where no masks are released, SAM~3.1 also segments the objects from prompts derived from the task description.
RAFT~\cite{teed2020raftrecurrentallpairsfield} flow between frames four steps apart, combined with the DA3 depth at both ends, gives the displacement of every arm and object pixel.
Pixels that move by less than 0.5\,px are set to zero motion, which suppresses depth flicker on static surfaces.
Actions are the end-effector positions, 6D rotations, and gripper states of both arms, normalized to $[-1,1]$ with dataset statistics.

\section{Simulation Experiments}
\label{sec:supp_sim}

We detail geometric supervision, post-training, and evaluation on RoboTwin~2.0, VLABench, and RoboDojo.

\subsection{Data and Geometric Supervision}
\label{sec:supp_sim_data}
All benchmarks label pixels as background, embodiment, or object, and foreground \trk{} stores the source-camera displacement $(\Delta u,\Delta v,\Delta d)$ to the next sampled frame.

\paragraph{RoboTwin 2.0.}
We replay each released seed and trajectory in SAPIEN, requiring joint states to match within $10^{-5}$.
Each frame records head-camera depth, actor segmentation, and all robot-link and scene-entity poses.
Robot links are labeled embodiment, and task objects are labeled object together with any other entity whose center moves more than 2\,mm.
Pixels are back-projected with simulator depth, transformed by the relative pose of their entity, and reprojected into the source camera.

\paragraph{VLABench.}
Official episodes supply depth and robot masks but lack object segmentation and poses.
We therefore regenerate 6,575 demonstrations with the official scene sampler, expert skill sequences, and success filter, changing only the recorder to save primary-camera depth, segmentation, and body poses per frame, and train on them together with the official data.
Franka arm and gripper bodies are labeled embodiment, and other bodies that translate more than 2\,mm or rotate more than $1^\circ$ are labeled object.
The rigid construction follows RoboTwin 2.0, with a median deviation of 0.8\,mm between transported and next-frame rendered depth.

\paragraph{RoboDojo.}
Released episodes contain head- and two-wrist-camera JPEG streams, per-frame intrinsics and extrinsics, joint states, end-effector poses, and actions, but no depth, segmentation, object poses, or scene seeds.
Using the released joint states and ARX X5 URDF, we render head-view robot depth and link indices in SAPIEN and derive embodiment \trk{} from the forward-kinematic motion of each link.
Remaining pixels use DA3 depth, SAM~3.1 object masks, and RAFT flow to the next sampled frame.
We align the episode-dependent depth scale of DA3 to the rendered robot depth $d^{\mathrm{R}}$ with one factor per episode,
\begin{equation}
  s=\operatorname{median}_{(t,p)\in\mathcal{R}}\;d^{\mathrm{R}}_t(p)\,/\,d^{\mathrm{DA3}}_t(p),
\end{equation}
where $\mathcal{R}$ collects the rendered robot pixels of all frames.
A shared scale avoids the fluctuations of per-frame fits, while a constant offset cancels in $\Delta d$.
Pixels inside the object masks whose flow exceeds 1\,px are labeled object and take $(\Delta u,\Delta v)$ from the flow and $\Delta d=s\,\big(d^{\mathrm{DA3}}_{t+4}(p')-d^{\mathrm{DA3}}_t(p)\big)$, where $p'$ is the flow correspondence of $p$.
The other non-robot pixels are treated as static background.

\paragraph{Current-frame conditions.}
The \trk{} expert receives the current head-view depth and a binary foreground mask, which covers the robot and task objects in RoboTwin 2.0 and the robot alone in VLABench.
Both benchmarks use simulator-rendered conditions during training and evaluation.
RoboDojo uses the robot-only renders above, recomputed at test time from live joint states and camera parameters rather than evaluation-simulator depth or segmentation, which keeps training and test inputs consistent.

\subsection{Training and Evaluation Protocols}
\label{sec:supp_sim_train}

\paragraph{Post-training.}
For $40\%$ of training samples, we draw $u\sim\mathcal{U}(0,1)$ and set $\sigma_{\mathrm{V}}=1-u$ and $\sigma_{\mathrm{G}}=\sigma_{\mathrm{A}}=\max(1-2u,0)$.
Another $40\%$ draw noise levels uniformly within $\sigma_{\mathrm{G}}=\sigma_{\mathrm{A}}\le\sigma_{\mathrm{V}}$.
The remaining $20\%$ keep \trk{} and actions clean with $\sigma_{\mathrm{V}}\sim\mathcal{U}(0,1)$.

\paragraph{Evaluation.}
Each RoboTwin 2.0 task is evaluated on 100 episodes per setting.
Following the official VLABench protocol, we run 50 episodes per task and track, except 10 for Insert Flower on the unseen-category track.
RoboDojo reports success rate and a partial-credit Score, each macro-averaged over its five evaluation dimensions.

\section{Real-Robot Experiments}
\label{sec:supp_real_setup}

We evaluate five tasks under training conditions and controlled visual shifts.

\subsection{Platform and Tasks}
\label{sec:supp_real_hardware}

\begin{figure}[t]
  \centering
  \includegraphics[width=0.88\linewidth]{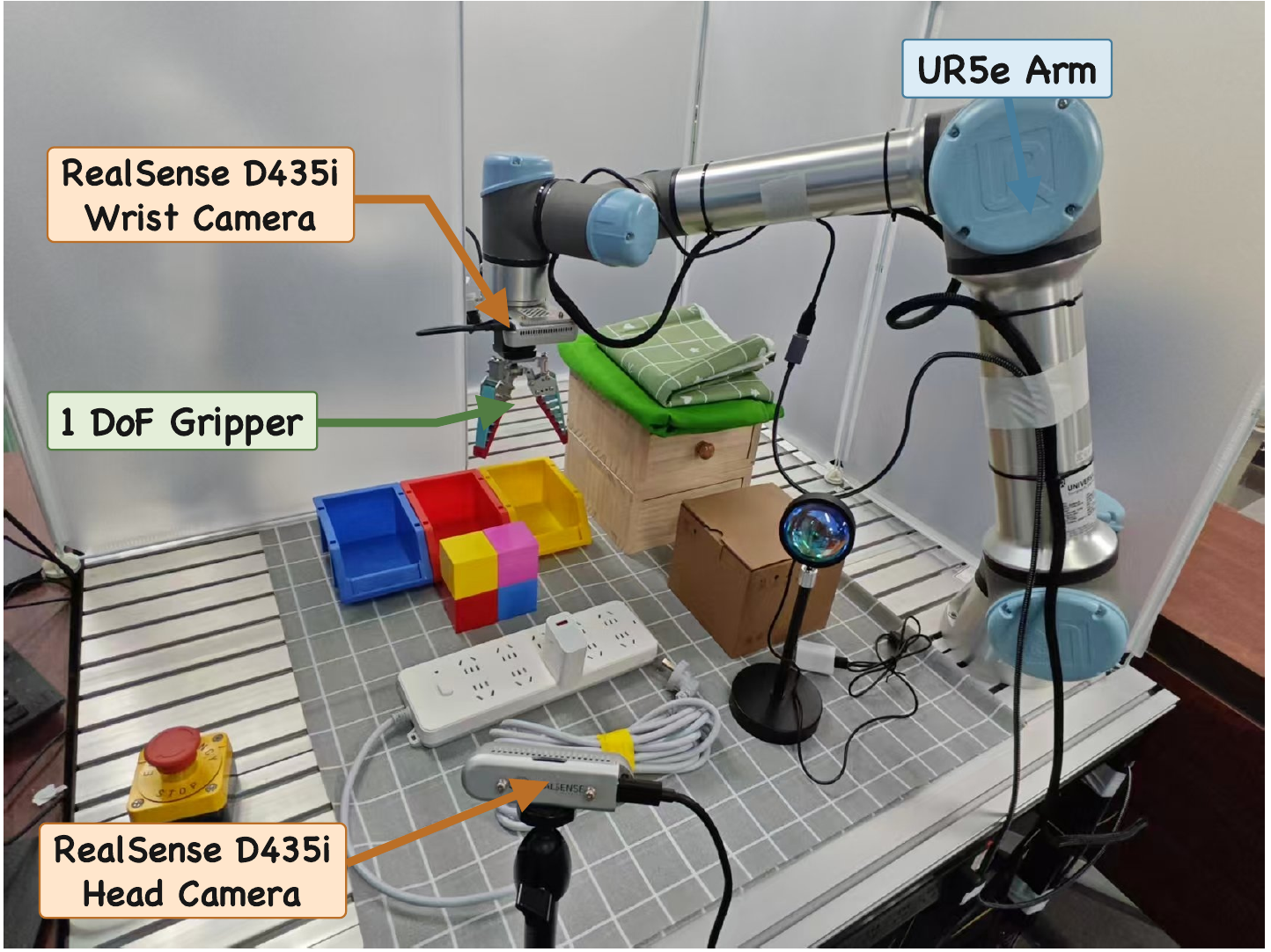}
  \caption{Real-robot platform. A UR5e arm with a 1-DoF parallel gripper operates in a tabletop workspace, observed by a fixed RealSense D435i head camera and a RealSense D435i wrist camera.}
  \label{fig:real_setup}
\end{figure}

The platform consists of a UR5e arm, a 1-DoF parallel gripper, and two RealSense D435i cameras (Fig.~\ref{fig:real_setup}, Table~\ref{tab:real_hardware}).
The head and wrist views provide RGB observations, and head-camera depth supplies the geometry for \trk{}.

\begin{table}[!tbp]
\centering
\caption{Real-robot acquisition and control settings. The 10\,Hz action grid specifies the nominal spacing of predicted targets, and replanning also includes inference time.}
\label{tab:real_hardware}
\footnotesize
\setlength{\tabcolsep}{4pt}
\renewcommand{\arraystretch}{1.1}
\begin{tabular}{@{}p{0.25\linewidth}p{0.69\linewidth}@{}}
\toprule
\textbf{Component} & \textbf{Configuration} \\
\midrule
Arm and gripper & Six arm joints and one gripper command, with measured joint states read through RTDE. \\
Camera streams & Head and wrist RGB-D at $640{\times}480$ and 30\,Hz, with frames saved at 10\,Hz. \\
Depth & Aligned to RGB with the camera SDK and converted to meters with the device depth scale. Zero marks invalid depth. \\
Action chunk & 32 targets on a 10\,Hz grid, of which the first 16 are executed before replanning. \\
Execution & Joint targets interpolated at 500\,Hz, with the arm holding its position during inference. \\
\bottomrule
\end{tabular}
\end{table}

\begin{figure*}[!tbp]
  \centering
  \includegraphics[width=0.90\textwidth]{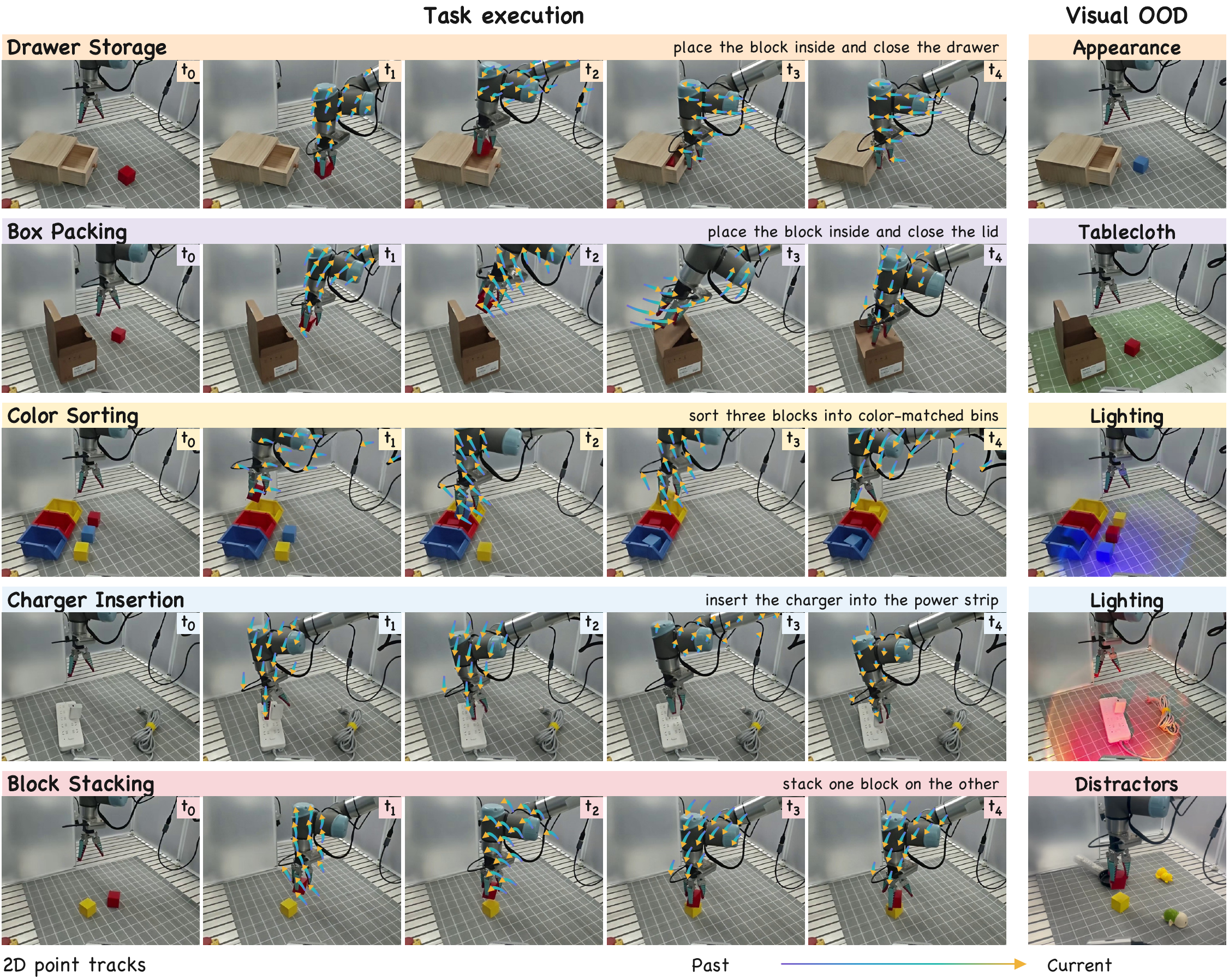}
  \caption{Real-robot tasks and visual OOD conditions. Each row is labeled with its task instruction, and the first five frames show one execution under training conditions. On frames $t_1$--$t_4$, gradient arrows show dense RAFT point tracks of the motion leading to each frame, with color progressing from purple (past) to gold (current) and arrowheads marking current positions. The right column shows separate trials with changed object appearance, tablecloth, lighting, or added distractors.}
  \label{fig:real_tasks}
\end{figure*}

\paragraph{Tasks.}
\label{sec:supp_real_tasks}
\begin{table}[!tbp]
\centering
\caption{Real-robot tasks and complete-task success criteria. Task instructions are shown in Fig.~\ref{fig:real_tasks}.}
\label{tab:real_task_criteria}
\footnotesize
\setlength{\tabcolsep}{4pt}
\renewcommand{\arraystretch}{1.1}
\begin{tabular}{@{}p{0.27\linewidth}p{0.67\linewidth}@{}}
\toprule
\textbf{Task} & \textbf{Success criterion} \\
\midrule
\multicolumn{2}{@{}l}{\textit{Routine}} \\
\quad Drawer Storage & The block is inside the drawer and the drawer is closed. \\
\quad Box Packing & The block is inside the box and the lid is closed. \\
\quad Block Stacking & The released red block stays on the yellow block. \\
\midrule
\multicolumn{2}{@{}l}{\textit{Long-horizon}} \\
\quad Color Sorting & The red, yellow, and blue blocks are each inside the bin of the same color. \\
\midrule
\multicolumn{2}{@{}l}{\textit{Precision}} \\
\quad Charger Insertion & The charger is fully inserted into the power strip. \\
\bottomrule
\end{tabular}
\end{table}

Table~\ref{tab:real_task_criteria} defines the five tasks and their success criteria.
Success requires completing the full task without human intervention, and intermediate progress receives no partial credit.
Figure~\ref{fig:real_tasks} illustrates the task sequences and representative visual shifts.

\subsection{Demonstrations and Geometric Supervision}
\label{sec:supp_real_data}

\begin{figure*}[!tbp]
  \centering
  \includegraphics[width=0.90\textwidth]{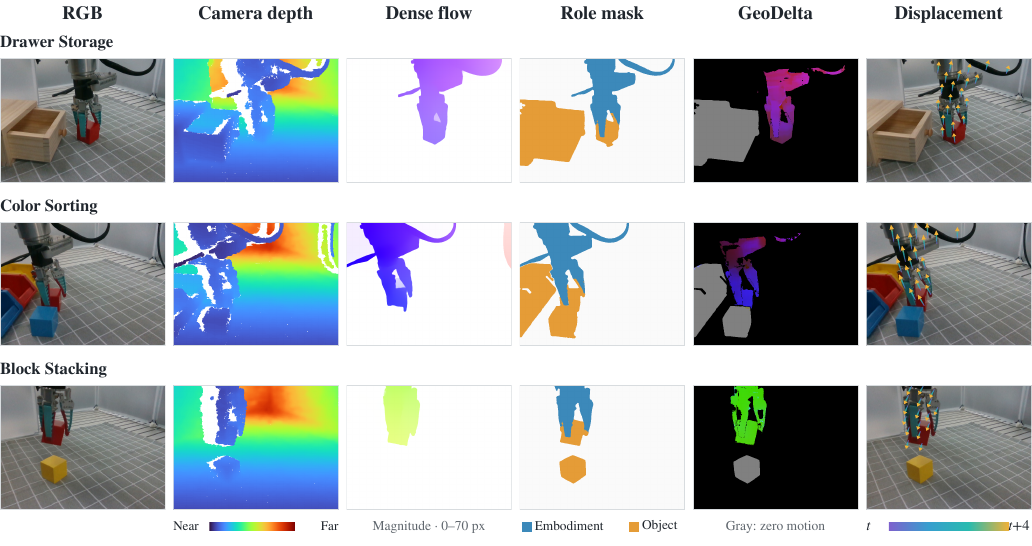}
  \caption{Geometric supervision on Drawer Storage, Color Sorting, and Block Stacking. From left to right: RGB, recorded camera depth, smoothed dense RAFT flow over the full image with linear magnitude scaling, SAM~3.1 role masks after morphological cleanup, RGB-encoded \trk{}, and displacement overlaid on RGB. All motion panels use the same frame pair, four control steps apart. Gradient arrows indicate image-plane displacement from $t$ to $t{+}4$.}
  \label{fig:real_processing_examples}
\end{figure*}

\paragraph{Demonstrations.}
We collect 50 robot demonstrations per task, 250 in total, and all methods are trained on the same data.
An operator guides the arm in freedrive mode and opens or closes the gripper with keyboard commands.
Object and movable-container placements vary within a $30{\times}30$\,cm region, while lighting, tablecloth, and object appearance stay fixed.
The two camera views are saved on a common 10\,Hz grid, and image timestamps are matched to the nearest recorded robot state.
Each training window contains the current observation and 32 future control steps, with video and \trk{} sampled every four steps.

\paragraph{Foreground and role masks.}
SAM~3.1~\cite{carion2026sam3segmentconcepts} segments the head-camera video with task prompts and optional boxes, and missing or drifting masks are corrected offline.
Pixels are labeled background, embodiment (arm and gripper), or object (manipulated objects and task containers).
The role labels supervise the interaction readout, and their foreground union conditions the \trk{} expert.
At deployment, the mask is computed from the current observation.

\paragraph{Geometric motion.}
For head-camera frames $t$ and $t{+}4$, RAFT-Large~\cite{teed2020raftrecurrentallpairsfield} estimates forward and backward flow with 24 updates.
At each source pixel, the forward flow gives $(\Delta u,\Delta v)$, and the depth change $\Delta d$ is the bilinearly sampled endpoint depth minus the source depth.
Together, they form \trk{} in the fixed head-camera frame.
A foreground correspondence is valid when its endpoint lies inside the image, both depths are valid, the forward--backward error is below 1\,px, and the embodiment or object role agrees across frames.
We apply the signed $\mu$-law displacement codec of the main paper with $\omega_u=\omega_v=W/6$, $\omega_d=0.10$\,m, and $\mu=31$.
Soft thresholds of 0.5\,px for image displacement and 2\,mm for depth change suppress label noise.
Figure~\ref{fig:real_processing_examples} shows the resulting supervision across tasks.

\begin{table}[!t]
\centering
\caption{Visual OOD factors. Each factor is evaluated in 10 trials per task, totaling 50 trials across the five tasks.}
\label{tab:real_ood_conditions}
\footnotesize
\setlength{\tabcolsep}{4pt}
\renewcommand{\arraystretch}{1.1}
\begin{tabular}{@{}p{0.22\linewidth}p{0.72\linewidth}@{}}
\toprule
\textbf{Factor} & \textbf{Setting} \\
\midrule
Lighting & Colored side illumination changes the light on the objects and workspace. \\
Tablecloth & The training tablecloth is replaced with an unseen green plaid tablecloth. \\
Appearance & Task objects have unseen appearances, including block-color changes in Drawer Storage and Box Packing. \\
Distractors & Non-target objects are added to the workspace. \\
\bottomrule
\end{tabular}
\end{table}

\begin{table}[!t]
\centering
\caption{Real-robot complete-task success counts. Each task has 40 ID and 40 OOD trials per method, and Total covers 200 trials per condition. \ours{} is best on every task in both conditions.}
\label{tab:real}
\footnotesize
\setlength{\tabcolsep}{1.85pt}
\renewcommand{\arraystretch}{1.08}
\begin{tabular}{@{}l*{8}{c}@{}}
\toprule
 & \multicolumn{2}{c}{\textbf{Fast-WAM}} & \multicolumn{2}{c}{\textbf{X-WAM}} & \multicolumn{2}{c}{\textbf{FlowWAM}} & \multicolumn{2}{c@{}}{\textbf{\ours{}}} \\
 & \multicolumn{2}{c}{\scriptsize RGB} & \multicolumn{2}{c}{\scriptsize RGB-D} & \multicolumn{2}{c}{\scriptsize RGB + 2D Flow} & \multicolumn{2}{c@{}}{\scriptsize RGB + \trk{}} \\
\cmidrule(lr){2-3}\cmidrule(lr){4-5}\cmidrule(lr){6-7}\cmidrule(l){8-9}
\textbf{Task} & \textbf{ID} & \textbf{OOD} & \textbf{ID} & \textbf{OOD} & \textbf{ID} & \textbf{OOD} & \textbf{ID} & \textbf{OOD} \\
\midrule
Drawer Storage    & 32 & 26 & 33 & 31 & 34 & 32 & \ourscol{\textbf{36}} & \ourslast{\textbf{35}} \\
Box Packing       & 24 & 20 & 28 & 25 & 30 & 26 & \ourscol{\textbf{33}} & \ourslast{\textbf{31}} \\
Block Stacking    & 29 & 22 & 30 & 26 & 33 & 24 & \ourscol{\textbf{35}} & \ourslast{\textbf{31}} \\
Color Sorting     & 21 & 16 & 23 & 20 & 29 & 21 & \ourscol{\textbf{32}} & \ourslast{\textbf{30}} \\
Charger Insertion & 11 & 10 & 23 & 22 & 25 & 19 & \ourscol{\textbf{29}} & \ourslast{\textbf{29}} \\
\midrule
Total             & 117 & 94 & 137 & 124 & 151 & 122 & \ourscol{\textbf{165}} & \ourslast{\textbf{156}} \\
\bottomrule
\end{tabular}
\end{table}

\begin{table}[!t]
\centering
\caption{Real-robot success counts under each visual OOD shift. Each shift aggregates 50 trials across the five tasks, and Total aggregates the 200 OOD trials of Table~\ref{tab:real}.}
\label{tab:real_ood}
\footnotesize
\setlength{\tabcolsep}{3.5pt}
\renewcommand{\arraystretch}{1.05}
\resizebox{\linewidth}{!}{%
\begin{tabular}{@{}lccccc@{}}
\toprule
\textbf{Method} & \textbf{Lighting} & \textbf{Cloth} & \textbf{Appearance} & \textbf{Distractors} & \textbf{Total} \\
\midrule
Fast-WAM~\cite{yuan2026fastwamworldactionmodels} & 24/50 & 29/50 & 18/50 & 23/50 & 94/200 \\
X-WAM~\cite{guo2026unified4dworldaction} & 31/50 & 35/50 & 27/50 & 31/50 & 124/200 \\
FlowWAM~\cite{chen2026flowwamopticalflowunified} & 31/50 & 38/50 & 25/50 & 28/50 & 122/200 \\
\midrule
\rowcolor{cvprblue!10}
\multicolumn{1}{@{}>{\columncolor{cvprblue!10}[0pt][\tabcolsep]}l}{\textbf{\ours{}}} & \textbf{39/50} & \textbf{41/50} & \textbf{36/50} & \textbf{40/50} & \multicolumn{1}{>{\columncolor{cvprblue!10}[\tabcolsep][0pt]}c@{}}{\textbf{156/200}} \\
\bottomrule
\end{tabular}}
\end{table}

\subsection{Training and Inference}
\label{sec:supp_real_training}

\paragraph{Action representation.}
At future step $\tau$, the action is $(\mathbf{q}_\tau-\mathbf{q}_0,g_\tau)$, where $\mathbf{q}_0$ contains the six current joint angles in radians and the absolute gripper command $g_\tau$ is 0 for open and 1 for closed.
Proprioception contains $\mathbf{q}_0$ and the current gripper command.
Joint actions and states are standardized with training-set statistics, and gripper commands are mapped to $[-1,1]$.
After inverse normalization, each joint displacement is added to $\mathbf{q}_0$ and executed with interpolation and velocity limits.

\paragraph{Post-training.}
\ours{} is post-trained for 5k steps on the 250 demonstrations after pretraining on \humanego{} and mid-training on Robot-30H.
We train one policy across all five tasks, conditioned on the task instruction.
Fast-WAM, X-WAM, and FlowWAM are fine-tuned from their released weights on the same demonstrations, with shared RGB views, action normalization, and execution settings.
X-WAM uses the aligned sensor depth, and FlowWAM constructs flow supervision with its public implementation.

\paragraph{Inference.}
\ours{} follows the coupled 20-step denoising schedule, with \trk{} and actions ready after step 10.
The controller executes the first 16 of the 32 predicted actions, then acquires a new observation and replans.
The remaining denoising steps refine Video only when a visual prediction is requested.

\subsection{Evaluation and Results}
\label{sec:supp_real_protocol}
\label{sec:supp_real_results}

Each method runs 40 ID and 40 OOD trials per task (400 in total), with methods alternating on shared initial configurations.
ID trials vary object and container placements within the training range.
OOD trials change one visual factor at a time (Table~\ref{tab:real_ood_conditions}), with 10 trials per factor and task.
The goal and all other conditions stay fixed, and appearance shifts preserve the block--bin color correspondence in Color Sorting.
Tables~\ref{tab:real} and~\ref{tab:real_ood} report complete-task success counts by task and by visual shift, using the criteria in Table~\ref{tab:real_task_criteria}.
These counts underlie Fig.~\ref{fig:real_results} of the main paper: \ours{} completes 165 of the 200 ID trials and 156 of the 200 OOD trials, corresponding to 82.5\% and 78.0\% SR.

\end{document}